\documentclass[journal]{IEEEtran}
\usepackage[authoryear]{natbib}

\usepackage{graphicx}  
\usepackage{svg}
\usepackage{amsmath}  
\usepackage{amssymb,amsfonts,latexsym}
\usepackage{enumerate}
\usepackage{xspace}
\usepackage{epsf,picinpar}
\usepackage{varioref}
\usepackage{colortbl,multirow,hhline}
\usepackage{listings}
\usepackage[center]{caption}
\usepackage[normalem]{ulem}
\usepackage{pifont}
\usepackage{soul}

\usepackage{url}
\usepackage{balance}
\usepackage{longtable}
\usepackage{lscape}
\usepackage{multirow}
\usepackage{listings}
\usepackage{framed}
\usepackage{morefloats}
\usepackage[T1]{fontenc}
\usepackage{array}
\usepackage{pdfpages}
\usepackage{fancybox}
\usepackage{flushend}
\usepackage{booktabs}
\usepackage{enumitem}
\usepackage{svg}
\usepackage{hyperref} 

\usepackage[table,xcdraw]{xcolor}
\usepackage{pdflscape}
\usepackage{tabularx}
\usepackage{lipsum}         
\usepackage{afterpage}      
\usepackage{float}
\usepackage{booktabs}

\usepackage{enumerate}
\usepackage{xspace}
\usepackage{epsf,picinpar}
\usepackage{varioref}
\usepackage{colortbl,multirow,hhline}
\usepackage{listings}
\usepackage[normalem]{ulem}

\usepackage{pifont}
\usepackage{xcolor,colortbl}

\usepackage[table,xcdraw]{xcolor}
\usepackage{pdflscape}
\usepackage{tabularx}
\usepackage{graphicx}

\usepackage{amssymb}
\usepackage{lineno}
\usepackage{pdflscape}  
\usepackage{float}

\usepackage{algorithm}
\usepackage{amssymb}
\usepackage{algorithmic}

\usepackage{xurl}

\usepackage{tcolorbox}
\usepackage{changepage} 
\tcbuselibrary{skins}


\usepackage[compact]{titlesec}
\usepackage {setspace}

\ifCLASSINFOpdf
\else
\fi

\begin{document}

\title{FedLLM: A Privacy-Preserving Federated Large Language Model for Explainable Traffic Flow Prediction}

\author{
Seerat~Kaur, Sukhjit~Singh~Sehra*, Dariush Ebrahimi
\\ 
Department of Physics \& Computer Science, Wilfrid Laurier University, Waterloo, Canada\\
\thanks{Email addresses: kaur8500@mylaurier.ca (Seerat Kaur), ssehra@wlu.ca (Sukhjit Singh Sehra*), debrahimi@wlu.ca (Dariush Ebrahimi)
}
\thanks{ORCID: Seerat Kaur (0009-0006-5627-1784), Sukhjit Singh Sehra (0000-0001-9058-7869), Dariush Ebrahimi (0000-0003-2489-8858)}
}


\maketitle

\begin{abstract}
Traffic prediction plays a central role in intelligent transportation systems (ITS) by supporting real-time decision-making, congestion management, and long-term planning. However, many existing approaches face several practical limitations. Most spatio-temporal models are trained in centralized settings, rely on numerical representations, and offer limited explainability. Recent Large Language Model (LLM)-based methods improve reasoning capabilities, but typically assume centralized data availability and do not fully capture the distributed and heterogeneous nature of real-world traffic systems. 
To address these challenges, this study proposes FedLLM, a privacy-preserving and distributed framework for explainable multi-horizon short-term traffic flow prediction (15–60 minutes ahead). The framework introduces four key contributions: (1) a Composite Selection Score (CSS) for data-driven freeway selection that captures structural diversity across traffic regions; (2) a domain-adapted LLM fine-tuned on structured traffic prompts encoding spatial, temporal, and statistical context; (3) an integrated federated learning (FL) and LLM framework that enables collaborative training across heterogeneous clients while exchanging only lightweight LoRA adapter parameters; and (4) a structured prompt representation that supports contextual reasoning and cross-region generalization. 
The FedLLM design allows each client to learn from local traffic patterns while contributing to a shared global model through efficient parameter exchange, reducing communication overhead and keeping data private. This setup supports learning under non-IID traffic distributions. Experimental results show that FedLLM achieves improved predictive performance over centralized baselines, while producing structured and explainable outputs. These findings highlight the potential of combining FL with domain-adapted LLMs for scalable, privacy-aware, and explainable traffic prediction.
\end{abstract}



\section{INTRODUCTION}

The rapid growth of urban populations, coupled with increasing vehicle densities, has intensified pressure on transportation networks, driving the need for traffic forecasting systems that are accurate, scalable, and practically interpretable. Within intelligent transportation systems (ITS), traffic prediction (also referred to as traffic forecasting) plays a central role by enabling real-time operational decisions, supporting proactive congestion management, and informing long-term infrastructure planning~\citep{Qu2024, Moghadas2024}.
Accurately modelling traffic flow remains a challenging task due to its dependence on non-linear dynamics, spatial variability across road networks, and temporally evolving patterns. These factors make it difficult for models to generalize consistently across different environments, time periods, and traffic conditions. Beyond predictive accuracy, real-world deployment increasingly requires models to provide interpretable outputs. Several studies have emphasized the importance of interpretability in traffic forecasting, highlighting that traffic operators must understand the reasoning behind a prediction to support time-critical decisions such as rerouting, signal adjustment, and incident response, a requirement that most existing models do not meet \citep{Kaur2026a, Gruver2023}. Graph-based spatio-temporal deep learning (DL) architectures have significantly advanced traffic forecasting performance \citep{Yu2018, Li2018, Bai2020, Guo2022}. These models capture spatial dependencies through graph convolution and temporal dynamics through recurrent or attention-based mechanisms, achieving strong results on established benchmarks. Despite this progress, they operate purely on numerical inputs and must infer contextual information, such as congestion patterns, demand trends, and temporal semantics, implicitly from the data. These models are trained in centralized settings and produce only numerical outputs, limiting their practical utility in scenarios where data sharing is restricted and interpretability is required.

Large language models (LLMs) have recently emerged as powerful tools for understanding, generating, and reasoning over structured and unstructured data \citep{Touvron2023, OpenAI2024, deepseek, Grattafiori2024, Qwen2025}. Although originally developed for natural language processing (NLP), their ability to encode contextual relationships, follow instructions, and produce interpretable outputs makes them increasingly relevant for time-series prediction and event explanation in urban mobility settings. In ITS, LLMs hold promise for traffic flow forecasting, incident explanation, accident prediction, and multimodal reasoning ~\citep{Moghadas2024, Ren2024, Li2024, Zhang2020, Jin2023, Guo2024, Zarza2023, Qu2024, Jin2021}. Their instruction-following capabilities, ability to integrate information from multiple sources, and generation of human-interpretable outputs are particularly valuable in systems requiring transparency and explainability. 
Despite their potential, deploying LLMs in ITS introduces several practical challenges. Firstly, most existing LLMs are pretrained on centralized corpora and assume access to large, labelled datasets, which contradicts the data sovereignty and privacy constraints commonly encountered in urban transportation infrastructures~\citep{Qi2023, Yurdem2024, Agarwal2023, Liu2023, Kaur2026}. Secondly, these models are resource-intensive and require significant memory and compute power, posing limitations for deployment on edge devices such as roadside units or smart sensors~\citep{Liu2024}. Without domain-specific fine-tuning, LLMs trained on generic corpora often underperform on specialized tasks such as traffic prediction~\citep{Kaur2026a}.
Another notable gap is the lack of spatio-temporal generalization. Most recently developed models and architectures, such as~\citep{Jin2023, Ren2024, Jin2021, Gruver2023}, primarily focus on univariate or multivariate time-series forecasting. While these models showcase temporal reasoning capabilities, they often ignore the spatial structure of traffic networks. As a result, they are not readily adaptable to scenarios where spatial heterogeneity and cross-region generalization are critical and hence limit their applicability in real-world ITS settings.

Federated Learning (FL) offers a promising framework to mitigate some of these challenges. FL allows multiple clients to collaboratively train and fine-tune a shared model without sharing raw data. This ensures privacy preservation and data sovereignty ~\citep{Qi2023}. By keeping data local and only sharing model updates (e.g., gradients or weight differentials), FL inherently supports data privacy, security, and regulatory compliance. It is important to note that here a client represents an independent data-holding entity such as a regional transportation authority, freeway management unit, or geographically distinct sensor network. In ITS, this is particularly beneficial given the widespread deployment of sensors across jurisdictions, each generating sensitive or proprietary data. In addition to privacy and scalability, FL enhances fault tolerance and system resilience. If a single node fails or opts out due to maintenance or connectivity issues, the remaining clients can continue training without compromising the integrity of the overall model. This distributed learning architecture is more resilient than centralized systems, which can suffer significant performance degradation or complete failure if a critical data source becomes unavailable. FL has also shown considerable utility for traffic flow prediction across geographically distributed and heterogeneous environments \citep{Liu2023, Yuan2023, Qi2023, Bonawitz2019, Kaur2026}. 
While FL is beneficial for decentralized model training and privacy preservation, it faces several limitations in the context of ITS. FL models rely on frequent communication between clients and the central server to exchange model updates, which can introduce significant communication overhead and latency, especially when dealing with high-dimensional traffic data and limited network bandwidth. These delays not only affect real-time prediction capabilities but also increase the risk of synchronization issues among clients, which reduces the overall efficiency of the system \citep{Kairouz2021, Bonawitz2019}. Another major limitation includes difficulty in capturing inter-client spatial dependencies \citep{Yang2024}, as traffic data is often fragmented across multiple owners. This fragmentation restricts interactions across clients. Moreover, FL-based systems remain vulnerable to privacy threats, such as inference attacks, which can extract sensitive information from local models or gradients. These vulnerabilities highlight the need for robust privacy-preserving computation protocols to ensure secure and efficient FL deployments. 

Combining LLMs with FL offers a unified response to both sets of limitations. Parameter-efficient fine-tuning methods such as LoRA \citep{Hu2021} and QLoRA \citep{Dettmers2023} make it feasible to adapt large models on resource-constrained hardware while exchanging only lightweight adapter parameters during aggregation, thus substantially reducing communication 
overhead. By representing traffic observations as structured natural language prompts, the model directly incorporates contextual information such as sensor metadata, spatial relationships, demand statistics, and temporal patterns into the prediction process. Another gap in existing work is the absence of a principled method for selecting training corridors. Most studies rely on arbitrary or geographic 
partitioning rather than data-driven criteria that reflect structural diversity across traffic environments. 
While both LLM-based forecasting and FL have been studied independently, their integration for explainable and privacy-preserving traffic flow prediction remains relatively 
underexplored in the literature \citep{Kaur2026a}. To our knowledge, no prior study has examined the use of federated LLM frameworks in the context of freeway traffic flow prediction. This study addresses these gaps through four 
contributions, evaluated on the California PeMS 
LargeST benchmark \citep{Liu2023a} using the Greater Los Angeles Area. The contributions of this study are described as follows:

\begin{enumerate}
\item \textbf{Composite Selection Score (CSS):} 
Existing freeway selection approaches typically 
rely on single statistics such as mean flow, 
which fail to capture the full structural 
diversity needed for representative model 
training. This study introduces the CSS, a 
novel multi-criteria scoring metric that 
simultaneously accounts for traffic volume, 
temporal variability, sensor reliability, and 
spatial coverage. Applied within a 
clustering-based selection pipeline, CSS ensures 
that training corridors represent structurally 
distinct traffic regimes, improving both domain 
adaptation quality and federated client 
diversity.

\item \textbf{Domain-Adapted LLM:} A 
Qwen2.5-1.5B-Instruct model~\citep{Qwen2025} is fine-tuned on 
structured traffic prompts using QLoRA, with 
each prompt encoding sensor metadata, long-run 
demand statistics, spatial neighbour dynamics, 
and temporal context. The model produces 
interval-wise flow predictions at 15, 30, 45, 
and 60 minutes alongside step-by-step natural 
language explanations grounded in the observed 
sensor context. The model consistently outperforms dedicated graph-based architectures, thus demonstrating strong data efficiency and generalization.

\item \textbf{FedLLM Framework:} To the best of our knowledge, this is the first federated LLM (FedLLM) framework applied to freeway traffic flow prediction. Four structurally diverse freeway corridors act as independent clients, each fine-tuning the domain-adapted model locally on private traffic 
prompts without sharing raw observations. Only lightweight LoRA adapter parameters are exchanged with a central server for aggregation, preserving data privacy and substantially 
reducing per-round communication cost relative to full model weight exchange. The FedLLM achieves stronger predictive performance than its centralized counterpart, demonstrating that training across heterogeneous client environments consistently improves generalization under non-IID conditions.

\item \textbf{Structured Prompt Representation for cross-region transfer:} A rigorous prompt design is developed to provide the base language model with structured, information-dense context at each prediction step. Each prompt encodes static sensor attributes, geographic coordinates, lane configuration, and graph-level connectivity metrics alongside long-run statistical summaries including mean flow, standard deviation (std), congestion ratio, and 24-hour and 7-day demand profiles. Dynamic temporal context comprising the most recent 12 flow observations, short-term trend slope, net flow change, and an interpretable time-of-day descriptor are also passed. Spatial context is further introduced through the recent flows of up to three same-freeway neighbouring sensors, enabling the model to account for upstream and downstream traffic propagation. This representation equips the model with the contextual grounding needed to reason about traffic conditions at a specific sensor and time, rather than relying solely on numerical extrapolation.

\end{enumerate}

The remainder of this paper is organized as follows. Section~\ref{related_work} surveys existing work on spatio-temporal traffic forecasting, LLM-based ITS applications, and FL for traffic prediction, positioning the proposed framework within the current literature. Section~\ref{methodology} details the proposed methodology, covering freeway selection via CSS, prompt generation, domain-adapted LLM training, and the FedLLM federated architecture. Section~\ref{Results_section} presents experimental results, baseline comparisons, ablation studies, and zero-shot evaluation. Section~\ref{limitation_FutureWork} discusses current limitations and directions for future work, and Section~\ref{Conclusion} concludes the paper with a summary of findings and broader implications.
\section{RELATED WORK}
\label{related_work}
Traffic prediction has evolved considerably over the past two decades. Early approaches relied on statistical methods such as 
AutoRegressive Integrated Moving Average (ARIMA) \citep{kumar2015}, which captured temporal patterns but struggled under the non-linear and non-stationary conditions typical of urban road networks. Classical machine learning (ML) methods, including support vector regression (SVR) \citep{Agarap2018} and Bayesian networks \citep{Wang2012}, offered greater flexibility but still treated spatial context as fixed input rather than a learnable component. DL approaches marked a significant step forward. \citet{Sutskever2014} introduced sequence-to-sequence learning with Fully connected Long Short-Term Memory (FC-LSTM) for modelling temporal dependencies in traffic sequences, while \citet{Cho2014} proposed the Gated Recurrent Unit (GRU) as a more computationally efficient recurrent alternative. Graph-based architectures then incorporated road network topology directly into the learning process. \citet{Yu2018} proposed a Spatio-temporal Graph Convolutional network (STGCN), combining graph and temporal convolution for joint spatio-temporal modelling. \citet{Li2018} introduced Diffusion Convolutional Recurrent Neural Network (DCRNN), treating traffic propagation as a diffusion process on a directed road graph with encoder-decoder recurrence for multi-step forecasting. \citet{Bai2020} proposed 
Adaptive Graph Convolutional Recurrent Network (AGCRN), learning node-specific graph structure adaptively rather than from a predefined adjacency matrix. \citet{Guo2022} introduced the Attention-based Spatio-Temporal Graph Neural Network (ASTGNN), incorporating dynamic spatial attention to capture evolving inter-sensor dependencies across prediction horizons.While these architectures collectively represent the state of the art in data-driven spatio-temporal forecasting and have demonstrated strong 
performance across established benchmarks, two practical constraints limit their broader deployment. First, they are typically trained under centralized data settings, which becomes 
problematic when traffic observations are distributed across independent agencies operating under distinct privacy constraints. Second, they operate on numerical sequences and 
produce only numerical outputs, without providing interpretable reasoning that traffic operators can examine to support operational decisions. These two constraints directly motivate the framework proposed in this study.

\begin{table*}[htbp]
\centering
\scriptsize
\caption{Evolution of FL in ITS: Contributions, Capabilities, Limitations, Strengths and Future directions}
\resizebox{\textwidth}{!}{
\begin{tabular}{p{2cm}p{0.5cm}p{3.2cm}p{3.8cm}p{2.8cm}p{3.5cm}p{2.8cm}p{3.5cm}}
\toprule
\textbf{Paper Title} & \textbf{Year} & \textbf{FL Use Case} & \textbf{Key Contributions} & \textbf{Capabilities} & \textbf{Limitations} & \textbf{Strengths} & \textbf{Future Directions} \\
\midrule
\citet{Yurdem2024} & 2024 & Overview of FL algorithms, tools, and domains & Surveys algorithms (FedAvg, FedYogi, FedProx, etc.), and tools like TensorFlow Federated, FedML, PySyft, Flower, and OpenFL across IoT, healthcare, finance and Traffic. & Broad FL foundation across privacy, architecture, application & No benchmarking or modeling,  Lacks quantitative performance evaluation & Clearly structured taxonomy of FL types (horizontal/vertical), strategies, applications, and tools. & Asynchronous FL, FLaaS (Federated Learning as a Service), blockchain-FL integration, defenses \\
\midrule
\citet{Nanayakkara2024} & 2024 & Survey of aggregation and optimization in FL, focusing on the theoretical gaps in FL aggregator design. & Provides taxonomy of aggregation rules, insights into client selection, fairness strategies, and robust aggregation (including statistical and performance-based) & Emphasizes robustness, personalization, and categorizes optimizers by heterogeneity and system constraints. & Theoretical only, lacks empirical benchmarks or experiments & Aggregation taxonomy across resilience & Quantum FL, solutions to dimensional collapse, and defense mechanisms against adversaries and system heterogeneity. \\
\midrule
\citet{Lu2024} & 2024 & Empirical and theoretical analysis of non-IID data impact on FL performance & Quantifies the degradation in FL performance using Earth Mover's Distance (EMD) to measure statistical skew, proposes a mitigation strategy by introducing a small globally shared dataset among clients. & Explains how skewed local class distributions lead to model divergence and decreased test accuracy & Introduces data sharing (privacy trade-off) & Clear non-IID data effect study, Presents a foundational statistical analysis & Suggests integrating personalized learning techniques and privacy-preserving mechanisms to enhance FL under real-world data distributions \\ 
\midrule
\citet{Zhang2024} & 2024 & CC-FedAvg for handling computational heterogeneity across clients & Captures the idea that clients dynamically choose between local training or estimation & Reduces device-side computation load & No personalization; lacks bias correction for skewed data & Practical cost-aware training & Multi-objective client selection \\
\midrule
\citet{Feng2024} & 2024 & FL-STGNN for traffic flow with heterogeneity handling & Divides road network via DTW+K-means; personalized and meme models with attention-based STGNNs; MFWA for aggregation & Highlights privacy-preserving, spatio-temporal personalized learning & Complex to scale; relies on accurate similarity clustering & Tackles model + objective heterogeneity & Extend to more privacy-preserving personalization and dynamic routing \\
\midrule
\citet{Qi2023} & 2023 & FL with Asynchronous GCNs for traffic prediction & Proposes FedAGCN using attention-weighted async aggregation with GRU-based ST modeling & Handles asynchronous and client staleness & Limited datasets, no strong personalization & Robust to communication delays and model staleness & Add DP protocols, model personalization, adaptive topology \\
\midrule
\citet{Agarwal2023} & 2023 & Survey of FL in ITMS & Reviews FL in traffic sign recognition, congestion detection, routing, parking, and object detection & Privacy-preserving, real-time edge processing & No unifying architecture, lacks empirical evaluation & Broad ITS applicability with practical insights & Integrate GNNs, adaptivity, and cross-silo scalability \\
\midrule
\citet{Yuan2023} & 2023 & FedSTN graph FL for urban traffic & Combines RLCN, AMFN (with FedGAT and VFL), SCN (POI-aware), with encrypted param sharing & Captures spatial-temporal + semantic patterns, Ensures privacy with AHE and parameter-level VFL & Needs POI tagging, assumes powerful edge servers & Strong semantic + encrypted FL modeling & Dynamic graph evolution, lightweight adaptive edge FL  \\
\midrule
\citet{Liu2023} & 2023 &  Multilevel FL for regional traffic & 2-tier MFL with VGAE for dynamic graph + STGNN for prediction & Captures semantic, spatial, and temporal dependencies while reducing communication overhead & Lacks personalization; no attention modeling & Scalable edge-cloud hierarchy with dynamic graph learning & Attention-based personalization, privacy-preserving enhancements \\
\midrule
\citet{Reddi2020} & 2020 & Adaptive FL optimization (FedOpt) & Proposes FedAdam, FedYogi, FedAdagrad using adaptive server-side optimizers & Improves convergence under data heterogeneity with minimal client overhead & Reduced but not eliminated client drift under heterogeneity & Mathematically grounded optimizer suite & Cross-device tuning, convergence-aware scheduling \\
\midrule
\citet{Bonawitz2019} & 2019 & Scalable FL infrastructure for mobile devices & Production FL with Secure Aggregation, DP, and actor-based coordination & Deployed on android devices at scale & Requires synchronous rounds; depends on device availability & Robust, privacy-preserving and fault-tolerant FL architecture & Bias mitigation, dynamic reconnection scheduling, global fairness \\
\midrule
\citet{Karimireddy2019} & 2019 & SCAFFOLD with control variates & Corrects client drift via control variates; variance reduction for stable updates & Robust to non-IID data and client sampling & Client state retention and hyperparameter tuning overhead & Unbiased updates, faster convergence than FedAvg & Personalized FL, lightweight control variate schemes \\
\midrule
\citet{McMahan2016} & 2016 & FedAvg - foundational aggregation method & Local SGD on clients + weighted global averaging; introduced FL paradigm & Simple, communication-efficient for large-scale systems & Sensitive to client drift; poor non-IID convergence; no personalization & Widely scalable with minimal infra assumptions & Fair/weighted FedAvg, personalization, adaptive aggregation \\
\midrule
\end{tabular}}
\label{tab:fl_summary}
\end{table*}

\subsection{LLMs in ITS}
The application of LLMs to ITS has grown considerably in recent years. \citet{Jin2021} introduced TrafficBERT, demonstrating that BERT-based representations pre-trained on large-scale traffic data transfer effectively to downstream flow forecasting tasks. 
\citet{Gruver2023} showed that LLMs can serve as zero-shot time series forecasters by encoding numerical sequences as text without task-specific training. \citet{Jin2023} proposed Time-LLM, reprogramming LLMs for time-series tasks by mapping input patches to text tokens. \citet{Liu2024} introduced ST-LLM, demonstrating that LLMs serve as effective temporal learners 
when combined with structured spatio-temporal inputs. \citet{Ren2024} introduced TPLLM, augmenting a pre-trained LLM with graph-structured spatial context for traffic prediction. 
\citet{Li2024} proposed UrbanGPT, integrating a spatio-temporal dependency encoder with instruction-tuning for generalizable urban forecasting under data scarcity. \citet{Guo2024} demonstrated that LLMs can generate natural language explanations alongside numerical traffic forecasts, directly addressing the 
interpretability gap. \citet{Moghadas2024} proposed Strada-LLM, embedding road network structure into the language model input for graph-aware prediction. \citet{Qu2024} introduced TrafficGPT, addressing long-context traffic analysis and generation by overcoming token limitations in standard LLM architectures. \citet{Zarza2023} explored multimodal LLM frameworks for traffic accident forecasting, combining visual and linguistic inputs for 
autonomous driving applications. \citet{Zhang2020} examined DL approaches for traffic state estimation, providing foundational context for LLM-based extensions in traffic modelling. A common limitation across these studies is the assumption that labelled data is available centrally for training or fine-tuning, which 
restricts their applicability in privacy-sensitive deployment settings. \citet{Kaur2026a} examined this landscape through a systematic review of 129 publications, categorizing LLM-based approaches in ITS into prompt-driven, adaptation-based, and domain-adapted methods. The review further highlights interpretability and privacy-preserving deployment as two persistent challenges across all three categories.

\subsection{FL for Traffic Prediction}
FL has attracted growing interest in traffic forecasting as a mechanism for collaborative model training under data privacy constraints. Table~\ref{tab:fl_summary} summarizes the key 
FL frameworks discussed in this section, highlighting their contributions, capabilities, and limitations across privacy preservation, aggregation strategy, and scalability. 
\citet{McMahan2016a} introduced federated averaging (FedAvg), 
establishing the foundational client-server aggregation paradigm for distributed model training. \citet{Liu2023} proposed a multilevel FL framework leveraging hierarchical 
server-client communication to reduce latency and exploit inter-region correlations, though it did not incorporate personalized adaptation or model interpretability. \citet{Yuan2023} introduced FedSTN, combining a recurrent long-term capture network, an attentive mechanism federated network, and a semantic capture network for urban traffic prediction on edge infrastructure, though it relied on 
external semantic information and required significant edge-computing capabilities. \citet{Qi2023} developed FedAGCN, integrating asynchronous FL with graph convolutional networks (GCN) for spatio-temporal modelling, though its aggregation remained static and did not adapt to client-level traffic context or support interpretability. \citet{Li2025} proposed Fed-GDAN, a graph diffusion attention network under federated aggregation for privacy-preserving flow prediction. \citet{Zhang2021} introduced FedASTGCN, a topology-aware federated framework for distributed traffic speed forecasting.
On the algorithmic side, \citet{Reddi2020} proposed adaptive federated optimizers to improve convergence under data heterogeneity, and \citet{Karimireddy2019} introduced SCAFFOLD to address client drift under non-IID conditions. 
\citet{Zhang2024} developed CC-FedAvg, a computationally customized framework allowing clients to adapt their update contribution based on available resources, though it did 
not address statistical heterogeneity or support adaptive personalization. \citet{Lu2024} confirmed through a comprehensive survey that standard aggregation strategies degrade substantially when client data distributions diverge, a condition common in ITS where traffic patterns vary across corridors and regions. More recently, targeted aggregation strategies have emerged: \citet{Feng2024} 
proposed client clustering and selective update fusion for personalized prediction; \citet{Tang2023} introduced differential privacy mechanisms at the local training level to enhance data protection while maintaining predictive accuracy; \citet{Zhang2024a} proposed personalized FL for cross-city traffic prediction, providing each client with 
a locally tailored model; and \citet{Yang2024} identified inter-client spatial dependency as a structural challenge arising from traffic data fragmentation across independent owners. \citet{Bonawitz2019} demonstrated scalable FL 
deployment in production environments with secure aggregation and differential privacy, confirming the practical viability of privacy-preserving FL at scale. \citet{Yurdem2024} provided a broad survey of FL algorithms, tools, and domains, covering frameworks such as Flower, TensorFlow Federated, and PySyft, thus offering a useful taxonomy of FL strategies across ITS and related fields. \citet{Nanayakkara2024} further examined aggregation rules and client selection strategies, emphasizing robustness and fairness in heterogeneous FL settings. \citet{Agarwal2023} surveyed FL applications across ITS tasks, including traffic sign recognition, congestion detection, and object detection, highlighting the breadth of federated approaches in transportation. \citet{Kaur2026} demonstrated that dynamically weighting client contributions based on six traffic-specific metrics, including load intensity, temporal freshness, and flow variability, consistently improves convergence and predictive performance under non-IID distributions, representing one of the first traffic-aware aggregation strategies in the FL literature.

Despite these advances, two fundamental limitations persist across existing federated traffic prediction frameworks. First, exchanging full model weights at each communication round introduces substantial overhead \citep{Bonawitz2019, Kairouz2021}, restricting scalability as the number of clients grows. Second, all current federated approaches produce only numerical predictions without interpretable reasoning, limiting their operational utility in settings where decision support and transparency are required. These frameworks also remain vulnerable to inference attacks that can extract sensitive information from shared gradients \citep{Kairouz2021}, highlighting the need for more robust 
privacy-preserving designs. To the best of our knowledge, no prior work has combined FL with domain-adapted LLMs for freeway traffic flow prediction in a framework that jointly addresses privacy preservation, communication efficiency, and 
interpretability. This study aims to bridge that gap.


\section{METHODOLOGY}
\label{methodology}
This study proposes a prompt-driven framework that integrates FL and LLMs for explainable traffic flow prediction. The goal is to forecast short-term traffic across multiple horizons while generating structured natural-language explanations grounded in spatial, statistical, and temporal context. Given historical traffic observations from freeway sensors, the framework predicts flows for the next four horizons (15-60 minutes) and provides interpretable explanations describing the predicted traffic dynamics. At the core of the framework is a domain-adapted LLM fine-tuned on structured traffic prompts derived from sensor observations and network metadata. This model operates both as a standalone explainable prediction model in a centralized setting and as the local model architecture within the FL environment. The federated framework reflects real-world traffic deployments where different agencies or traffic management units exhibit heterogeneous traffic patterns and sensor coverage. To preserve data privacy, raw observations remain local to each client, while only model updates are shared with a central server. This section outlines the key components of the proposed
approach, including the problem formulation, data preparation and freeway selection using the CSS, prompt generation, domain-adapted LLM training, and the federated learning architecture. Performance is evaluated across multiple prediction horizons, varying test set sizes, and seasonal splits, with additional zero-shot cross-region experiments on unseen geographic areas to assess predictive accuracy, scalability, and generalization capability. 

\subsection{Problem Definition}
Short-term traffic flow prediction estimates how traffic conditions evolve in the near future based on recent observations and contextual knowledge of the road network. In this study, we predict freeway traffic flow across multiple time horizons while producing interpretable explanations alongside each forecast.
In practice, traffic sensors are typically administered by separate regional authorities, such as district transportation agencies, freeway management units, or municipal traffic control centres, each operating over a geographically distinct freeway segment. Due to ownership restrictions, infrastructure security concerns, and inter-agency governance constraints, these entities rarely share raw data across administrative boundaries. As a result, centralizing traffic data collected from multiple jurisdictions into a single training repository is neither feasible nor desirable. This creates a fundamental challenge: models that achieve strong predictive performance usually require large and diverse training data, yet the data needed to build such models cannot always be freely aggregated.
Let $x_t^{(i)}$ denote the traffic flow measured at sensor $i$ at time $t$, expressed in vehicles per 15-minute interval. For each sensor and timestamp, the model receives a sequence of the most recent $L$ observations:
\begin{equation}
\mathbf{x}_{t-L:t-1}^{(i)} =
\left\{
x_{t-L}^{(i)},\,
x_{t-L+1}^{(i)},\,
\ldots,\,
x_{t-1}^{(i)}
\right\}
\label{eq:input}
\end{equation}
where $L = 12$ corresponds to the previous three hours of traffic history. The goal is to forecast traffic flow for the next four intervals:
\begin{equation}
\hat{\mathbf{x}}_{t:t+3}^{(i)} =
\left\{
\hat{x}_t^{(i)},\,
\hat{x}_{t+1}^{(i)},\,
\hat{x}_{t+2}^{(i)},\,
\hat{x}_{t+3}^{(i)}
\right\}
\label{eq:output}
\end{equation}
representing predictions at 15, 30, 45, and 60 minutes ahead. The road network is represented as a directed weighted graph $\mathcal{G} = (\mathcal{V}, \mathcal{E}, \mathbf{A})$, where $\mathcal{V}$ is the set of $N$ loop detector stations, $\mathcal{E}$ is the set of directed road connections, and $\mathbf{A} \in \mathbb{R}^{N \times N}$ is the adjacency matrix, with entries 
derived from pairwise road network distances via a thresholded 
Gaussian kernel.
Most existing traffic forecasting models discussed in the literature rely only on numerical time-series data. In contrast, our study formulates traffic prediction as a structured reasoning task. Along with historical flow sequences, the proposed prediction models receive contextual information describing the surrounding traffic environment, including spatial relationships between sensors, long-run statistical summaries, and temporal context such as time of day and weekly demand patterns. The forecasting function is therefore written as:
\begin{equation}
\hat{\mathbf{x}}_{t:t+3}^{(i)} =
f_\theta
\left(
\mathbf{x}_{t-L:t-1}^{(i)},\,
\mathbf{c}_t^{(i)}
\right)
\label{eq:model}
\end{equation}
where $\mathbf{c}_t^{(i)}$ denotes the contextual information associated with sensor $i$ at time $t$, including sensor metadata, neighbouring traffic activity, long-run traffic statistics, and temporal descriptors.

To address the challenges of scalability and data privacy, we design two novel prediction models. The first is a centralized domain-adapted LLM, where traffic data from multiple freeways is aggregated to fine-tune a base LLM (Qwen) for explainable traffic forecasting. While this approach allows the model to learn from a larger unified dataset, it assumes that raw data can be freely shared across regional boundaries.
The second model is the FedLLM framework, representing a novel integration of FL and LLMs for traffic flow prediction. To the best of our knowledge, this is the first study to apply LLM-based modelling within a federated setting for spatio-temporal traffic prediction.
In this decentralized setting, each freeway acts as an independent client that trains the model locally using its own traffic data. Only model updates are exchanged with a coordinating server, while the underlying observations remain private within each client’s infrastructure. These clients represent structurally diverse traffic environments, ranging from low-volume peripheral routes to high-flow commuter expressways, reflecting the heterogeneous and non-IID nature of traffic data.
Both of these models are designed to produce interpretable forecasts. Alongside each prediction, the model generates a structured output that includes interval-wise flow estimates, concise analytical explanations describing the observed traffic conditions, step-by-step reasoning that links the forecast to spatial and temporal context, and a quantified trend label. This helps ensure that the forecasting process is transparent, constructive, and insightful across different traffic regimes.
By comparing the two models, the study examines how decentralized training can achieve competitive forecasting performance while improving scalability, supporting cross-region generalization, and preserving data privacy across independent clients.

\subsection{Dataset Preparation and Prompt Generation}
\label{data_preperation}
\begin{figure}[h]
  \centering
  \includegraphics[width=\columnwidth]{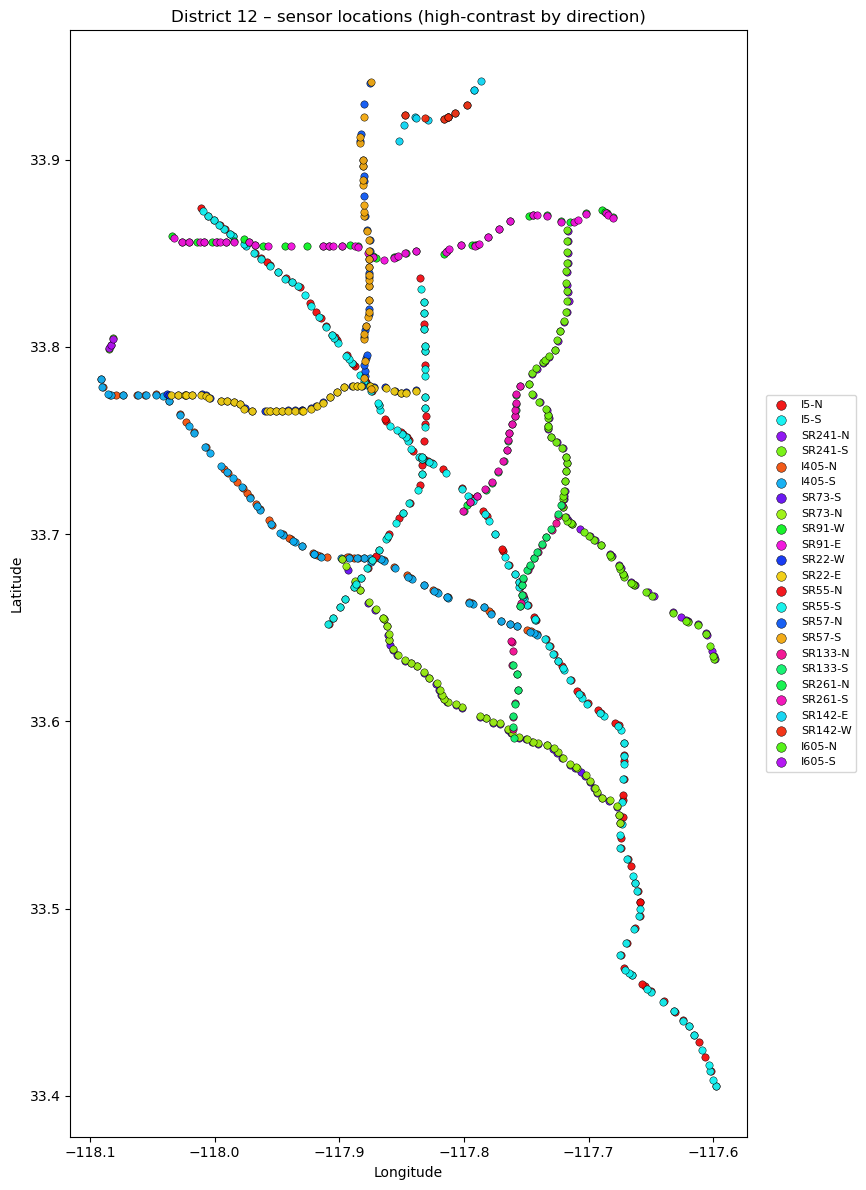}
  \caption{District 12 - Geographical distribution of traffic sensors in PeMS. Each colour represents a different freeway and travel direction, illustrating the spatial layout of sensors. }
  \label{fig:D12_AllHighways}
\end{figure}

This study utilizes the LargeST dataset \citep{Liu2023a}, a large-scale benchmark derived from the Caltrans Performance Measurement System (PeMS) \citep{CaltransWebsite}. The dataset spans five years (2017 to 2021) and covers 8,600 mainline loop detector stations across the California freeway network, originally recorded at 5-minute intervals and aggregated to 15-minute windows following \citep{Liu2023a}.
The road network is represented as a weighted directed graph with adjacency matrix $\tilde{\mathbf{A}} \in \mathbb{R}^{8600 \times 8600}$, where edge weights are derived from road network distances using a threshold Gaussian kernel. Each sensor record includes traffic flow readings alongside metadata covering geographic coordinates, district, highway identifier, travel direction, and lane count.
LargeST was selected because its extensive spatial coverage, detailed temporal granularity, and comprehensive sensor metadata make it ideal for developing and testing models that incorporate both spatial and contextual awareness. Figure~\ref{fig:D12_AllHighways} further illustrates the geographic distribution of 953 sensor nodes across 24 freeway-direction corridors in District 12, highlighting the spatial diversity of the network.
In our experiments, we focus on two regional subsets. District 12 of the Greater Los Angeles Area (GLA), covering Orange County, is utilized as the primary region for training and testing. Meanwhile, District 4 of the Greater Bay Area (GBA) is held out entirely for zero-shot cross-region generalization, with no overlap with the training pipeline.

\begin{figure*}[htbp]
\centering
\resizebox{0.74\textwidth}{!}{%
\begin{minipage}{\textwidth}
\includegraphics[width=0.48\textwidth]{ 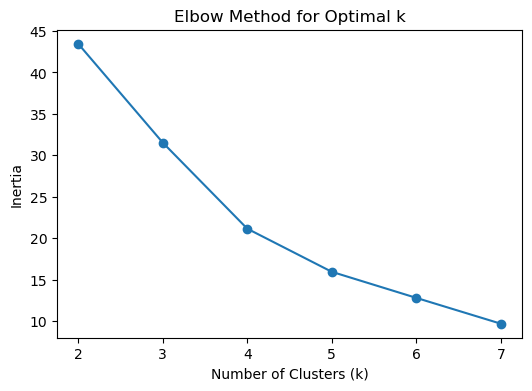}
\hfill
\includegraphics[width=0.48\textwidth]{ 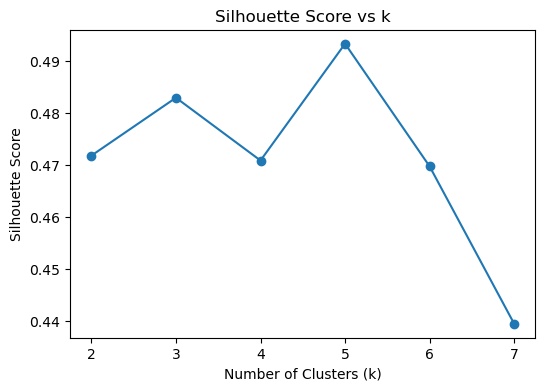}
\vspace{0.4cm}
\includegraphics[width=0.48\textwidth]{ 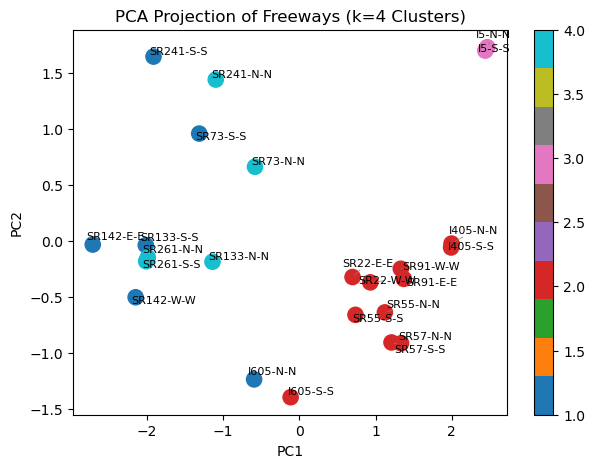}
\hfill
\includegraphics[width=0.48\textwidth]{ 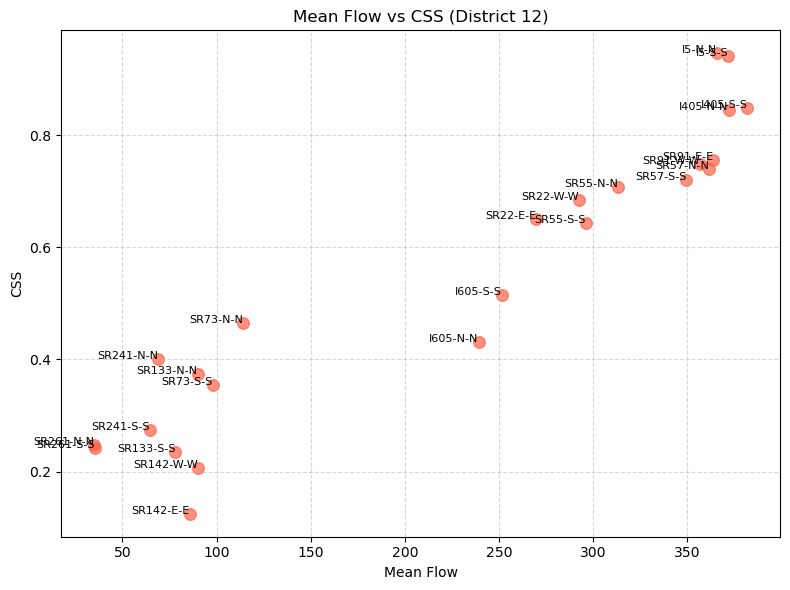}
\end{minipage}
}
\caption{Freeway selection criteria using clustering and the proposed CSS score. Panels show the elbow method for selecting cluster count (top-left), silhouette score for clustering quality (top-right), PCA projection displaying separation among freeway groups (bottom-left), and the relationship between mean traffic flow and CSS for ranking freeway suitability (bottom-right).}
\label{fig:css_selection}
\end{figure*}

\subsubsection{Freeway Characterization, Clustering, and CSS Ranking}
Rather than working with individual sensors directly, we treat each freeway-direction corridor as a single structural unit. All sensor observations within a corridor are pooled, and four descriptors are computed for corridor $i$: mean flow $\mu_i$, standard deviation (std) $\sigma_i$, zero-rate $\rho^{(0)}_i$, and sensor count $n_i$, forming the feature vector $\mathbf{f}_i = (\mu_i,\, \sigma_i,\, \rho^{(0)}_i,\, n_i)$. 
The mean flow captures the overall traffic demand intensity, standard deviation reflects the temporal variability across congestion and free-flow regimes, zero-rate measures the data sparsity which arises from sensor inactivity or low utilization, and sensor count represents the spatial coverage. Together, these four dimensions provide a structural representation of each freeway segment, allowing us to analyze the transition from raw sensor-level time series to corridor-level traffic descriptors suitable for clustering and ranked selection.
The 24 corridors are then clustered in the standardized feature space $\hat{\mathbf{F}} \in \mathbb{R}^{24 \times 4}$, where each column is z-score normalized so that no single dimension dominates the Euclidean distances. K-means clustering is then applied over $k \in \{2,\ldots,7\}$, where the optimal $k$ is determined using the elbow method and silhouette scoring combination. The elbow method evaluates the cluster inertia as follows:
\begin{equation}
    J(k) = \sum_{c=1}^{k} \sum_{i \in \mathcal{C}_c} 
    \|\hat{\mathbf{f}}_i - \boldsymbol{\mu}_c\|_2^2
    \label{eq:inertia}
\end{equation}
where $\mathcal{C}_c$ denotes the freeways in clusters $c$ and $\boldsymbol{\mu}_c$ is the cluster centroid. As seen in Figure~\ref{fig:css_selection} (top-left), inertia bends sharply at $k = 4$, after which gains diminish. The silhouette score peaks at $k = 5$ ($\approx 0.493$) but holds reasonably at $k = 4$ (top-right). Given the elbow result and the practical need for interpretable traffic clusters, $k = 4$ is used throughout the study. Figure~\ref{fig:css_selection} (bottom-left) further shows the PCA projection of the four-dimensional feature space, confirming that the four clusters are well-separated in structure. The four clusters thus identified correspond to distinct traffic regimes: high-capacity interstate backbone corridors, high-volume commuter expressways, moderate-flow secondary corridors, and low-volume peripheral routes. 
\begin{figure*}[h]
  \centering
  \includegraphics[width=0.75\textwidth, trim=0cm 0cm 0cm 2cm, clip]{ 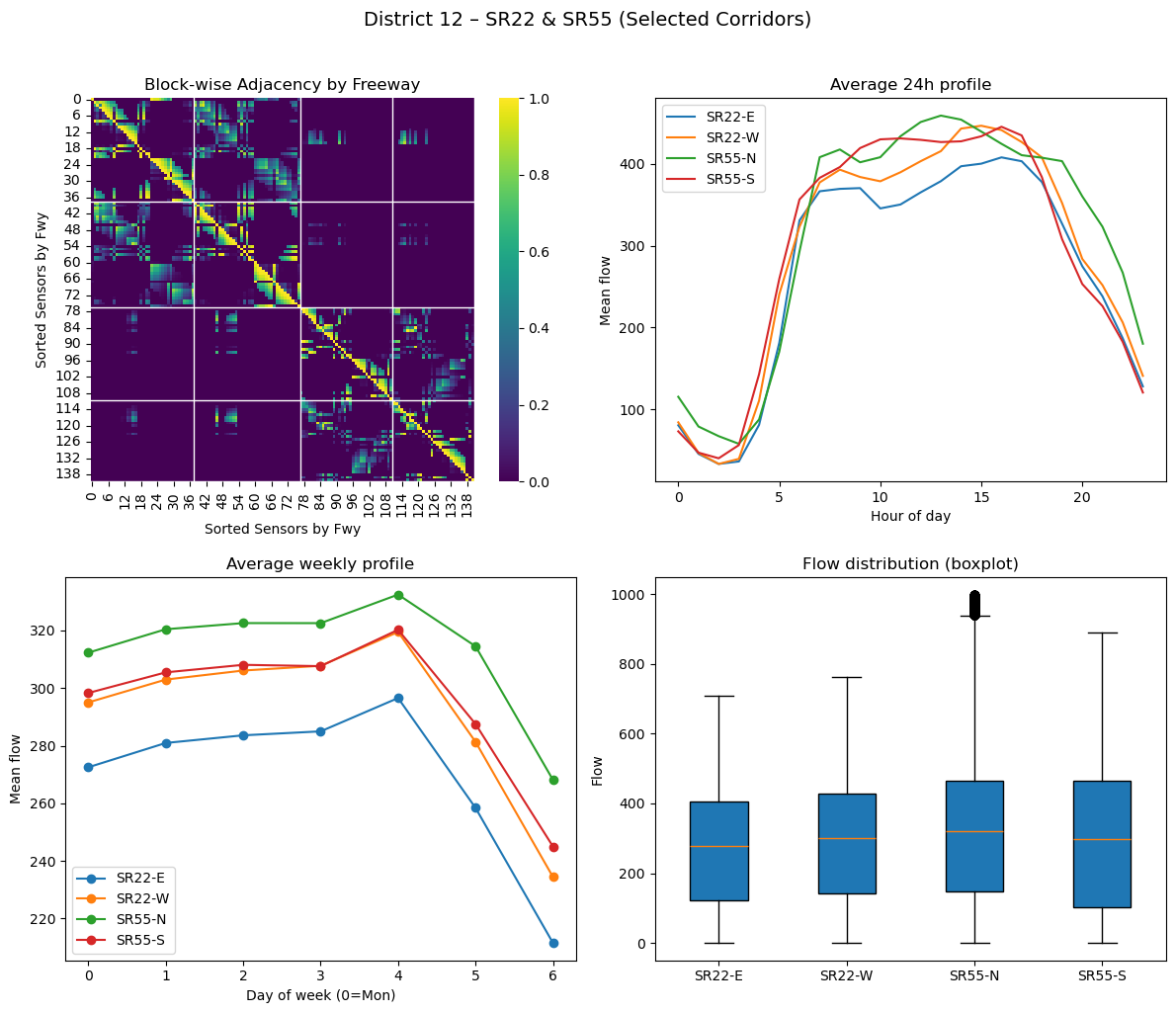}
  \caption[Exploratory analysis of the selected PeMS District 12 corridors used in this study
(SR22-E/W and SR55-N/S). ]{Exploratory analysis of the selected PeMS District 12 corridors used in this study (SR22-E/W and SR55-N/S). The figure shows (top-left) the adjacency submatrix representing spatial relationships among sensors, (top-right) the average 24-hour traffic flow profile, (bottom-left) the average weekly traffic pattern, and (bottom-right) the distribution of traffic flow across the selected corridors.}
  \label{fig:D12_QwenCentralizedTraining_SR22_SR55}
\end{figure*}
\begin{table}[ht]
\centering
\scriptsize
\caption{CSS rankings for District 12, sorted in descending order. Selected corridors for the domain-adapted LLM (SR22-E/W, SR55-N/S) and FL clients (SR261-S, SR57-N, SR133-N, SR133-S) are highlighted.}
\label{tab:css_rankings}
\begin{tabular}{clccccc}
\toprule
\textbf{Rank} & \textbf{Freeway} & \textbf{Sensors} & 
\textbf{Mean Flow} & \textbf{Std} & \textbf{Zero-rate} & \textbf{CSS} \\
\midrule
1  & I5-N     & 107 & 365.93 & 201.68 & 0.003 & 0.946 \\
2  & I5-S     & 106 & 371.78 & 200.15 & 0.007 & 0.942 \\
3  & I405-S   & 57  & 381.82 & 211.72 & 0.006 & 0.849 \\
4  & I405-N   & 58  & 371.95 & 231.39 & 0.032 & 0.845 \\
5  & SR91-E   & 44  & 363.63 & 171.69 & 0.000 & 0.756 \\
6  & SR91-W   & 46  & 356.70 & 181.65 & 0.023 & 0.748 \\
7  & \textbf{SR57-N}   & 29  & 361.40 & 189.07 & 0.002 & 0.740 \\
8  & SR57-S   & 28  & 349.40 & 183.38 & 0.002 & 0.721 \\
9  & \textbf{SR55-N}   & 34  & 313.31 & 190.24 & 0.014 & 0.708 \\
10 & \textbf{SR22-W}   & 39  & 292.49 & 172.15 & 0.010 & 0.684 \\
11 & \textbf{SR22-E}   & 38  & 269.84 & 161.06 & 0.011 & 0.650 \\
12 & \textbf{SR55-S}   & 30  & 296.08 & 203.28 & 0.078 & 0.643 \\
13 & I605-S   & 32  & 251.39 & 130.76 & 0.007 & 0.515 \\
14 & SR73-N   & 51  & 113.97 & 114.37 & 0.055 & 0.465 \\
15 & I605-N   & 32  & 39.19  & 177.12 & 0.155 & 0.431 \\
16 & SR241-N  & 66  & 68.89  & 71.39  & 0.066 & 0.401 \\
17 & \textbf{SR133-N}  & 24  & 89.91  & 79.90  & 0.015 & 0.375 \\
18 & SR73-S   & 52  & 98.02  & 108.71 & 0.157 & 0.354 \\
19 & SR241-S  & 64  & 64.72  & 74.49  & 0.199 & 0.273 \\
20 & SR261-N  & 17  & 34.97  & 44.26  & 0.040 & 0.247 \\
21 & \textbf{SR261-S}  & 16  & 35.45  & 51.18  & 0.053 & 0.242 \\
22 & \textbf{SR133-S}  & 20  & 78.09  & 86.83  & 0.154 & 0.236 \\
23 & SR142-W  & 7   & 90.14  & 101.88 & 0.183 & 0.206 \\
24 & SR142-E  & 14  & 85.68  & 90.51  & 0.269 & 0.125 \\
\bottomrule
\end{tabular}
\end{table}
\begin{figure*}[htbp]
\centering
\includegraphics[width=0.32\textwidth]{ 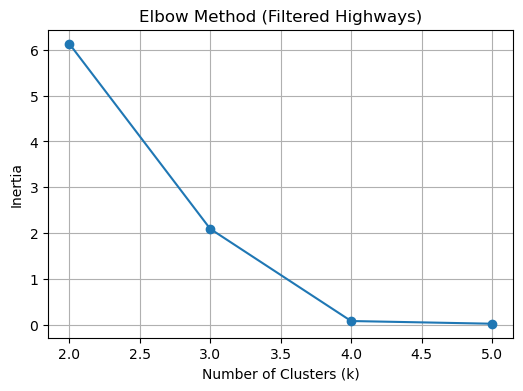}
\hfill
\includegraphics[width=0.32\textwidth]{ 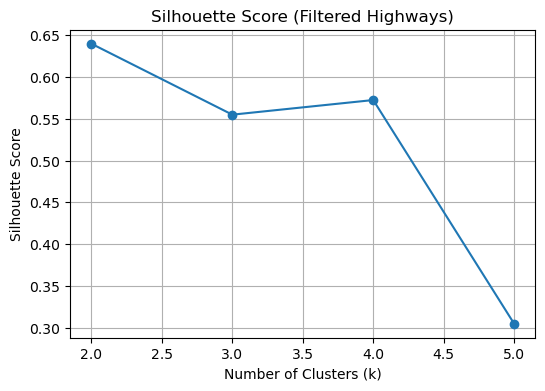}
\hfill
\includegraphics[width=0.29\textwidth]{ 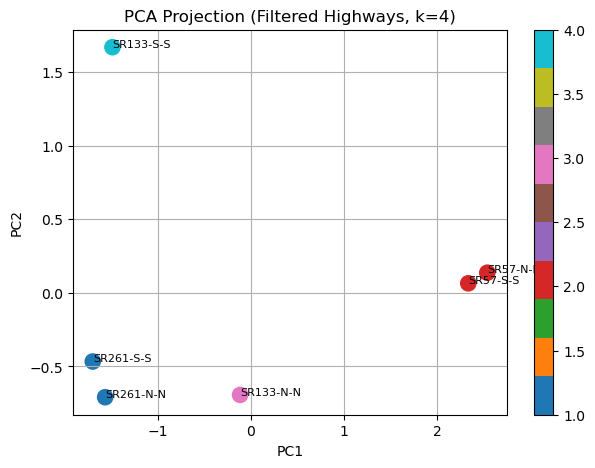}
\caption[Clustering diagnostics for the filtered freeway subset used for FL client construc-
tion. Freeway segments with 15–40 sensors ]{Clustering diagnostics for the filtered freeway subset used for FL client construction. Freeway segments with 15–40 sensors were re-clustered using structural features (mean flow, variability, zero-rate, and sensor count). The elbow (left) and silhouette (middle) analyses support a four-cluster solution, while the PCA projection (right) visualizes the resulting clusters used to select representative FL corridors.
}
\label{fig:filtered_clustering}
\end{figure*}
To rank corridors within and across these clusters, we introduce the \textbf{Composite Selection Score (CSS)}, a novel multi-criteria scoring metric for freeway selection. Rather than selecting freeways based on a single statistic such as mean flow, CSS simultaneously accounts for traffic flow/volume, temporal variability, sensor reliability, and spatial coverage. All features are min-max normalized to $[0,1]$, and the zero-rate term is inverted prior to normalization, so that corridors with consistently active sensors receive higher scores while penalizing those with high data sparsity.
\begin{equation}
    \text{CSS}_i = \alpha\,\hat{\mu}_i + \beta\,\hat{\sigma}_i 
    + \gamma\,(1 - \hat{\rho}^{(0)}_i) + \delta\,\hat{n}_i
    \label{eq:css}
\end{equation}
where $\hat{(\cdot)}$ denotes min-max normalization and $\alpha = \beta = \gamma = \delta = 0.25$ , summing up to one. Equal weights are used so each factor contributes equally, and the result is easy to reproduce and interpret.
Table~\ref{tab:css_rankings} reports the full CSS rankings for all freeways in District 12 alongside their constituent feature values. 
As shown in Figure~\ref{fig:css_selection} (bottom-right), CSS broadly correlates with mean flow yet assigns higher ranks to corridors that additionally exhibit strong temporal variability, low sparsity, and adequate sensor coverage, making it a more complete suitability criterion than mean flow alone.

\subsubsection{Freeway Selection and  Data Partitioning}
Using the cluster assignments and CSS rankings established above, freeway-direction corridors are selected independently for each of the following proposed models:

\textbf{Domain-Adapted LLM:} Based on the global clustering results, SR22-E, SR22-W, SR55-N, and SR55-S are selected for domain adaptation. All four freeways belong to the high-volume commuter cluster and carry CSS values between 0.643 and 0.708, as reported in Table~\ref{tab:css_rankings}. While I-5 and I-405 achieve the highest CSS scores ($\approx 0.94$), their sensor counts exceed 100 each, placing them beyond the scope of a controlled fine-tuning experiment. SR22 and SR55 offer a balanced alternative, collectively contributing 141 sensors that provide sufficient data density for effective domain adaptation. 
Figure~\ref{fig:D12_QwenCentralizedTraining_SR22_SR55} provides an exploratory analysis of the selected corridors. The adjacency submatrix (top-left) reveals intra-corridor connectivity, reflecting the underlying freeway topology. The 24-hour demand curves (top-right) show a gradual morning build-up followed by sustained afternoon peaks across all four corridors, confirming their structural consistency within the same traffic regime. The weekly profiles (bottom-left) exhibit a consistent weekday-to-weekend demand decline across all corridors, and the flow distributions (bottom-right) show comparable median flows and interquartile ranges, further validating that the four corridors form a coherent and homogeneous training set. We utilized the January through June 2019 subset for training and July through December for testing to ensure the model is exposed to a broad range of seasonal demand variation within a single calendar year. The GBA District 4 corridors SR24-E/W and SR87-N/S (107 sensors) are reserved exclusively for zero-shot cross-region generalization, where the model is evaluated on an entirely unseen geographic region without any retraining or fine-tuning.

\textbf{Federated Clients:} Based on the same clustering approach applied earlier, a secondary filtering step is applied to retain only freeways 
with 15 and 40 sensors, ensuring comparable local dataset sizes across FL clients. At this stage, SR22 and SR55 were excluded to prevent data leakage from the domain adaptation task. Six corridors remained after filtering: SR261-N, SR261-S, SR133-N, SR133-S, SR57-N, and SR57-S. K-means is then re-applied to this filtered subset using the same standardized feature space. As shown in Figure~\ref{fig:filtered_clustering}, the elbow curve 
(left) drops sharply to near-zero inertia by $k = 4$, and the 
silhouette scores (middle) support a well-defined partition at this 
value, with $k = 4$ selected to ensure structural diversity across 
clients, and the PCA projection (right) confirming clear structural 
separation among the four resulting clusters. The top CSS-ranked freeways from each cluster were then selected, yielding SR261-S (Client 1, low-volume, 16 sensors), SR57-N (Client 2, high-flow, 29 sensors), SR133-N (Client 3, moderate stable, 24 sensors), and SR133-S (Client 4, moderate sparse, 20 sensors), covering 89 sensors in total. Moreover, the GBA District 4 corridors SR24-E/W and SR87-N/S (107 sensors) are used for zero-shot evaluation of the FedLLM framework.

\begin{figure*}[htbp]
  \centering
  \includegraphics[width=0.8\textwidth]{ 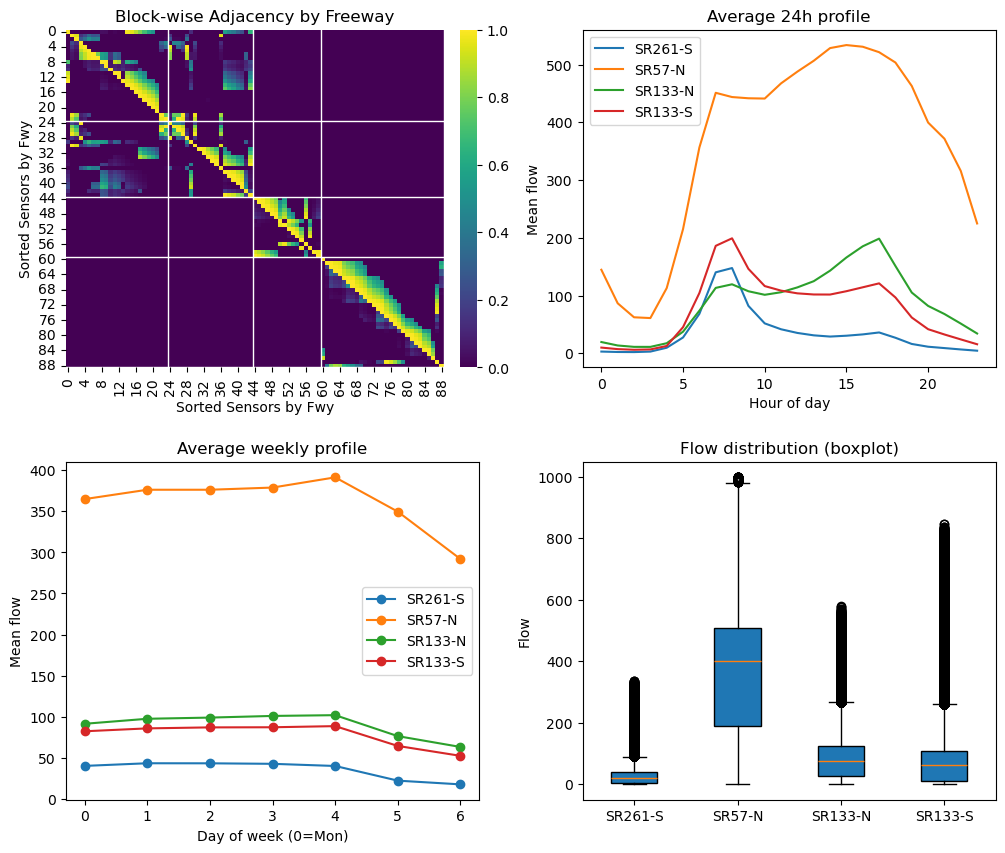}
  \caption[Exploratory analysis of the freeway corridors selected as FL clients (SR261-S, SR57-
N, SR133-N, and SR133-S).]{Exploratory analysis of the freeway corridors selected as FL clients (SR261-S, SR57-N, SR133-N, and SR133-S). The panels show the adjacency submatrix capturing spatial relationships among sensors( top-left), the average 24-hour traffic flow profile (top-right), the average weekly traffic pattern (bottom-left), and the distribution of traffic flow across the selected corridors (bottom-right).}
  \label{fig:D12_FLclients_profiles}
\end{figure*}

Figure~\ref{fig:D12_FLclients_profiles} further presents the exploratory analysis of the four selected clients. The adjacency submatrix (top-left) shows the spatial connectivity structure across the four selected corridors. The 24-hour profiles (top-right) reveal distinct demand levels across clients, 
capturing a broad range of traffic intensities from low-volume to high-flow conditions. The weekly profiles (bottom-left) show varying weekday and weekend patterns across clients, and the flow distributions (bottom-right) confirm that the four corridors occupy structurally different traffic regimes, ensuring heterogeneous and representative conditions for federated evaluation. The same chronological split is applied, where January-June 2019 is used for local client training and July-December for evaluation.

\subsubsection{Prompt Generation}
Following the dataset partitioning described above, each sensor observation is converted into a structured natural language prompt paired with a ground-truth JSON response, serving as the training and evaluation input for both proposed models. The same schema is applied across domain-adapted LLM training, FL client training, testing, and zero-shot evaluation. These meticulously designed prompts allow the model to operate on structured traffic knowledge while leveraging the reasoning capabilities of LLMs. Each training record consists of a system prompt defining the model role, a user prompt containing the structured sensor context, and the ground-truth JSON response.

Each prompt combines three categories of information. \textit{Static sensor context} includes the sensor ID, geographic coordinates, district, freeway identifier, travel direction, lane count, and detector type, alongside structural graph features derived from $\mathbf{A}$ such as in-degree and out-degree. 
\textit{Statistical context} summarizes the long-term behaviour of sensor $i$ using statistics computed from the training data, including mean flow $\bar{x}^{(i)}$, std $\sigma^{(i)}$, congestion ratio 
$\rho^{(i)}_{\text{cong}}$, observed flow range $[x^{(i)}_{\min}, x^{(i)}_{\max}]$, a 24-hour demand profile $\mathbf{h}^{(i)} \in \mathbb{R}^{24}$, and a 7-day weekly profile $\mathbf{w}^{(i)} \in \mathbb{R}^{7}$. \textit{Dynamic temporal context} is computed at each timestep $t$ and includes the historical sequence $\mathbf{x}_{t-L:t-1}^{(i)}$, 
short-term trend slope estimated from the most recent observations, net flow change over the last 30 minutes, the typical mean flow at the current clock hour, and an interpretable time-of-day 
descriptor such as morning peak (commute buildup) or evening peak (homeward commute).
Spatial context is introduced through the most recent flows from up to three same-freeway neighbours, selected by the largest adjacency weights in row $i$ of $\mathbf{A}$. This enables the prompt to reflect local traffic propagation effects while avoiding unrelated cross-freeway signals.
The ground-truth response is formatted as a structured JSON object containing predicted traffic flows at 15, 30, 45, and 60 minutes, average future flow, trend label, and trend change describing whether traffic is increasing, decreasing, stable or mixed. To support explainable forecasting, each response also includes three explanation components: an interval-wise explanation describing the expected change at each horizon, a concise analytical explanation summarizing the current traffic state, and a step-by-step explanation linking the prediction to temporal patterns, congestion level, and neighbouring sensor behaviour. All explanations are generated deterministically from the observed data, ensuring numerical consistency between the reasoning text and the predicted values.

\begin{figure*}[htbp]
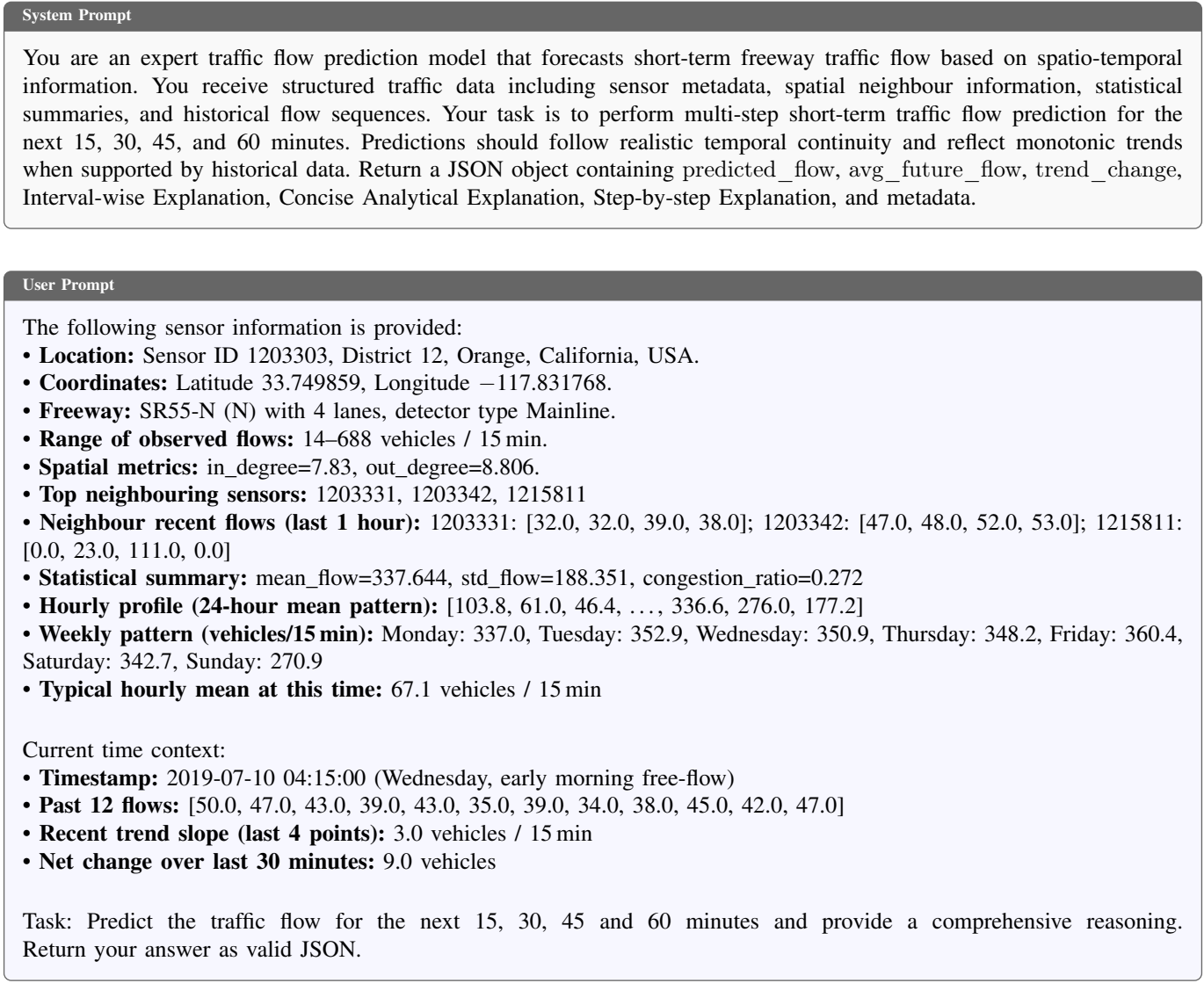

\centering
\begin{tcolorbox}[
    colback=gray!5,
    colframe=black!60,
    fonttitle=\bfseries\scriptsize,
    title=System Prompt,
    boxrule=0.5pt,
    left=5pt, right=5pt, top=5pt, bottom=5pt
]
{
You are an expert traffic flow prediction model that forecasts 
short-term freeway traffic flow based on spatio-temporal 
information. You receive structured traffic data including sensor 
metadata, spatial neighbour information, statistical summaries, 
and historical flow sequences. Your task is to perform multi-step 
short-term traffic flow prediction for the next 15, 30, 45, and 
60 minutes. Predictions should follow realistic temporal continuity 
and reflect monotonic trends when supported by historical data. 
Return a JSON object containing \texttt{predicted\_flow}, 
\texttt{avg\_future\_flow}, \texttt{trend\_change}, Interval-wise 
Explanation, Concise Analytical Explanation, Step-by-step 
Explanation, and metadata.
}
\end{tcolorbox}
\vspace{6pt}
\begin{tcolorbox}[
    colback=blue!3,
    colframe=black!60,
    fonttitle=\bfseries\scriptsize,
    title=User Prompt,
    boxrule=0.5pt,
    left=5pt, right=5pt, top=5pt, bottom=5pt
]
{
The following sensor information is provided:\\
\textbullet\ \textbf{Location:} Sensor ID 1203303, District 12, 
Orange, California, USA.\\
\textbullet\ \textbf{Coordinates:} Latitude 33.749859, 
Longitude $-$117.831768.\\
\textbullet\ \textbf{Freeway:} SR55-N (N) with 4 lanes, 
detector type Mainline.\\
\textbullet\ \textbf{Range of observed flows:} 
14--688 vehicles / 15\,min.\\
\textbullet\ \textbf{Spatial metrics:} 
in\_degree=7.83, out\_degree=8.806.\\
\textbullet\ \textbf{Top neighbouring sensors:} 
1203331, 1203342, 1215811\\
\textbullet\ \textbf{Neighbour recent flows (last 1 hour):} 
1203331: [32.0, 32.0, 39.0, 38.0]; 
1203342: [47.0, 48.0, 52.0, 53.0]; 
1215811: [0.0, 23.0, 111.0, 0.0]\\
\textbullet\ \textbf{Statistical summary:} 
mean\_flow=337.644, std\_flow=188.351, 
congestion\_ratio=0.272\\
\textbullet\ \textbf{Hourly profile (24-hour mean pattern):} 
[103.8, 61.0, 46.4, \ldots, 336.6, 276.0, 177.2]\\
\textbullet\ \textbf{Weekly pattern (vehicles/15\,min):} 
Monday: 337.0, Tuesday: 352.9, Wednesday: 350.9, 
Thursday: 348.2, Friday: 360.4, Saturday: 342.7, 
Sunday: 270.9\\
\textbullet\ \textbf{Typical hourly mean at this time:} 
67.1 vehicles / 15\,min\\[2pt] \newline
Current time context:\\
\textbullet\ \textbf{Timestamp:} 2019-07-10 04:15:00 
(Wednesday, early morning free-flow)\\
\textbullet\ \textbf{Past 12 flows:} [50.0, 47.0, 43.0, 39.0, 
43.0, 35.0, 39.0, 34.0, 38.0, 45.0, 42.0, 47.0]\\
\textbullet\ \textbf{Recent trend slope (last 4 points):} 
3.0 vehicles / 15\,min\\
\textbullet\ \textbf{Net change over last 30 minutes:} 
9.0 vehicles\\[2pt] \newline
Task: Predict the traffic flow for the next 15, 30, 45 and 60 minutes and provide a comprehensive reasoning. Return your answer as valid JSON.}

\end{tcolorbox}
\caption{Designed prompt structure for freeway traffic flow 
prediction (Part 1 of 2): system prompt and user prompt.}
\label{fig:prompt_example_1}
\end{figure*}

In preliminary experiments, we observed that the model occasionally produced near-zero or flat output sequences inconsistent with the observed trend - a known tendency of language models that exhibit simplicity and repetition biases when applied to numerical sequences \citep{Gruver2023}. Light adjustments are therefore applied to the raw target values prior to finalization, preventing near-zero predictions when traffic activity is clearly present and correcting flat 
forecasts when recent observations show a directional trend. All values are constrained within $[x^{(i)}_{\min}, x^{(i)}_{\max}]$ to ensure physically plausible outputs. The same schema is applied across centralized training, federated client training, and zero-shot evaluation, ensuring both paradigms receive identical contextual information and enabling a direct comparison between centralized and decentralized settings. 
Figure~\ref{fig:prompt_example_1} presents the designed prompt structure, including the system prompt and user prompt, while Figure~\ref{fig:prompt_example_2} illustrates a representative output produced by the FedLLM, demonstrating the structured JSON format and natural language reasoning chain generated for a sample traffic prediction.

\begin{figure*}[htbp]
\centering
\begin{tcolorbox}[
    colback=green!3,
    colframe=black!60,
    fonttitle=\bfseries\scriptsize,
    title=Model Output,
    boxrule=0.5pt,
    left=5pt, right=5pt, top=5pt, bottom=5pt
]
\{\\
``predicted\_flow'': [73.0, 98.0, 115.0, 137.0],\\
``avg\_future\_flow'': 105.75,\\
``trend\_change'': +90.0,\\
``trend\_label'': increasing,\\ \\
``\textbf{Interval-wise Explanation}'': \{\\
``15-min ahead'': ``Volume is climbing towards $\sim$73.0 (+26.0), indicating strengthening traffic conditions.",\\
``30-min ahead'': ``Flow is expected to rise to $\sim$98.0 (+51.0), thus reflecting a buildup along SR55-N.",\\
``45-min ahead'': ``Traffic projected to increase to $\sim$115.0 (+68.0) which is typical of rising morning peak.",\\
``60-min ahead'': ``Flow is expected to rise to $\sim$137.0 (+90.0), typical of rising morning peak (commute buildup) demand.",\\
``Overall'': ``The sequence exhibits an increasing trend based on recent dynamics (net change over last 45\,min = +9.0\,veh, CV=0.01, congestion ratio=0.27). Neighbour patterns from sensors 
1203331, 1203342, 1215811 and nearby segments show comparable early morning free-flow behaviour in Orange."\\
\},\\ \\
``\textbf{Concise Analytical Explanation}'': ``Observed trends suggest traffic is expected to increase based on projected short-term demand. Despite the early morning hours, the demand is beginning to build up. Nearby freeway segments show comparable short-term behaviour. [Congestion: moderate, ratio=0.272].",\\ \\
``\textbf{Step-by-step Explanation}'': [\\
``1. Time Context: 4:15 on Wednesday (early morning free-flow).",\\
``2. Historical Observation: Last 12 readings [50.0, 47.0, 43.0, 39.0, 43.0, 35.0, 39.0, 34.0, 38.0, 45.0, 42.0, 47.0], avg=41.8, slope=$-$0.3.'',\\
``3. Flow Regime: free-flow regime ($<$40\% of typical capacity).'',\\
``4. Congestion \& Variability: ratio=0.27 (moderate), historical std=188.4, CV=0.11.'',\\
``5. Spatial Context: Neighbouring sensors (1203331, 1203342, 1215811) suggest similar freeway behaviour.",\\
``6. Predicted Future Trend: Next 15-60\,min flows [73.0, 98.0, 115.0, 137.0], avg=105.75, delta(last)=+90.0.'',\\
``7. Interpretation: Demand is beginning to build in the early pre-commute window, consistent with the projected morning ramp-up pattern on SR55-N, despite a slightly negative short-term 
slope in recent observations. The prediction trends observed align with observed early morning free-flow dynamics.''\\
],\\ \\
``\textbf{metadata}": \{``sensor\_id'': ``1203303'', ``timestamp'': ``2019-07-10 04:15:00''\}\\
\}
\end{tcolorbox}
\caption[Output produced by the FedLLM for the input prompt in Figure 4.6, illustrating the
structured prediction format including interval-wise flow estimates,]{Output produced by the FedLLM for the input prompt in Figure~\ref{fig:prompt_example_1}, illustrating the structured prediction format including interval-wise flow estimates, a concise analytical summary, and a step-by-step reasoning chain 
grounded in the observed sensor context.}
\label{fig:prompt_example_2}
\end{figure*}

\subsection{Domain-Adapted LLM}
\label{domain_adapted_LLM}
Supervised fine-tuning is a widely adopted technique for adapting pre-trained LLMs to specialized domains \citep{Wei2021, Ouyang2022}. Although such models acquire broad linguistic and reasoning capabilities from large-scale general corpora, they typically require additional training on task-specific data in order to perform reliably in specialized application settings.
To enable domain adaptation for traffic forecasting, we employ Qwen2.5-1.5B-Instruct \citep{Qwen2025} as the base language model. This model was selected due to its strong instruction capabilities, efficient parameter scale, and demonstrated performance on structured reasoning tasks. With approximately 1.5 billion parameters, it provides a balance between expressive capability and computational efficiency, making it suitable for large-scale traffic datasets while remaining feasible for fine-tuning on a single GPU.
The model is loaded using the Unsloth framework \citep{unsloth}, which provides kernel-level optimizations for faster transformer training. To further reduce memory usage, the base model is initialized using 4-bit NormalFloat (NF4) quantization.

We formulate traffic forecasting as a prompt-driven, structured generation task. Each training sample contains a system prompt defining the model role, a user prompt $P_i$ describing the sensor context at time $t$, and a ground-truth assistant response containing the target predictions and explanations. During training, the loss is computed over the assistant response tokens so that the model learns to generate accurate predictions and explanations.
\begin{equation}
\mathcal{L}_{\text{SFT}} = -\frac{1}{|Y|} 
\sum_{j=1}^{|Y|} \log P_\theta\!\left(y_j \mid 
x_1, \ldots, x_T, y_1, \ldots, y_{j-1}\right)
\label{eq:sft_loss}
\end{equation}
where $x_1, \ldots, x_T$ are the tokenised input context tokens and $y_1, \ldots, y_{|Y|}$ are the target response tokens.
Full fine-tuning of a 1.5B-parameter model over more than 2.4 million prompts would be computationally expensive. We therefore employ QLoRA \citep{Dettmers2023}, which combines 4-bit quantization of frozen base weights with low-rank trainable adapters \citep{Hu2021}. Instead of updating the full weight matrices, each adapted layer learns a low-rank update:
\begin{equation}
\mathbf{W}' = \mathbf{W} + \Delta\mathbf{W} = \mathbf{W} + \mathbf{B}\mathbf{A}
\label{eq:lora}
\end{equation}
where $\mathbf{W} \in \mathbb{R}^{d \times k}$ denotes 
the frozen pre-trained weights, $\mathbf{A} \in 
\mathbb{R}^{r \times k}$ and $\mathbf{B} \in 
\mathbb{R}^{d \times r}$ are trainable low-rank matrices with rank $r = 16$, and $\mathbf{B}$ is initialized to zero, so the adapter contributes no perturbation at the 
start of training. LoRA adapters are inserted into the query, key, value, and output projections of each attention layer, as well as the gate, up, and down projections of the feed-forward layers. With a scaling factor $\frac{\alpha}{r}=1$, 1.18\% of the model parameters are trained. Despite updating only a small fraction of the total parameters, the model converges effectively and achieves stable and accurate forecasting performance, demonstrating that parameter-efficient adaptation is sufficient for this domain.
Training is conducted on a single NVIDIA A100-SXM4-80GB GPU. The dataset is randomly shuffled using seed 3407 to ensure the model observes traffic patterns across different months and time-of-day periods rather than learning from chronological sequences. Each record is converted into chat-formatted text using Qwen’s native chat template prior to tokenization so that the training format matches the inference setting.
Supervised fine-tuning is performed with an effective batch size of 16 (per-device batch size of 2 with gradient accumulation over 8 steps), a learning rate of $2\times10^{-4}$, a weight decay of 0.01 and 50 warm-up steps, training runs for 3000 optimization steps, fused AdamW optimizer and cosine learning-rate scheduling. Computation is performed in bfloat16 precision to maximize throughput while maintaining numerical stability. After training, the LoRA adapter weights are merged with the base model parameters to produce a standalone domain-adapted traffic forecasting model that can be used directly for inference and evaluation.
\begin{figure*}[t]
    \centering
    \includegraphics[width=\textwidth]{ 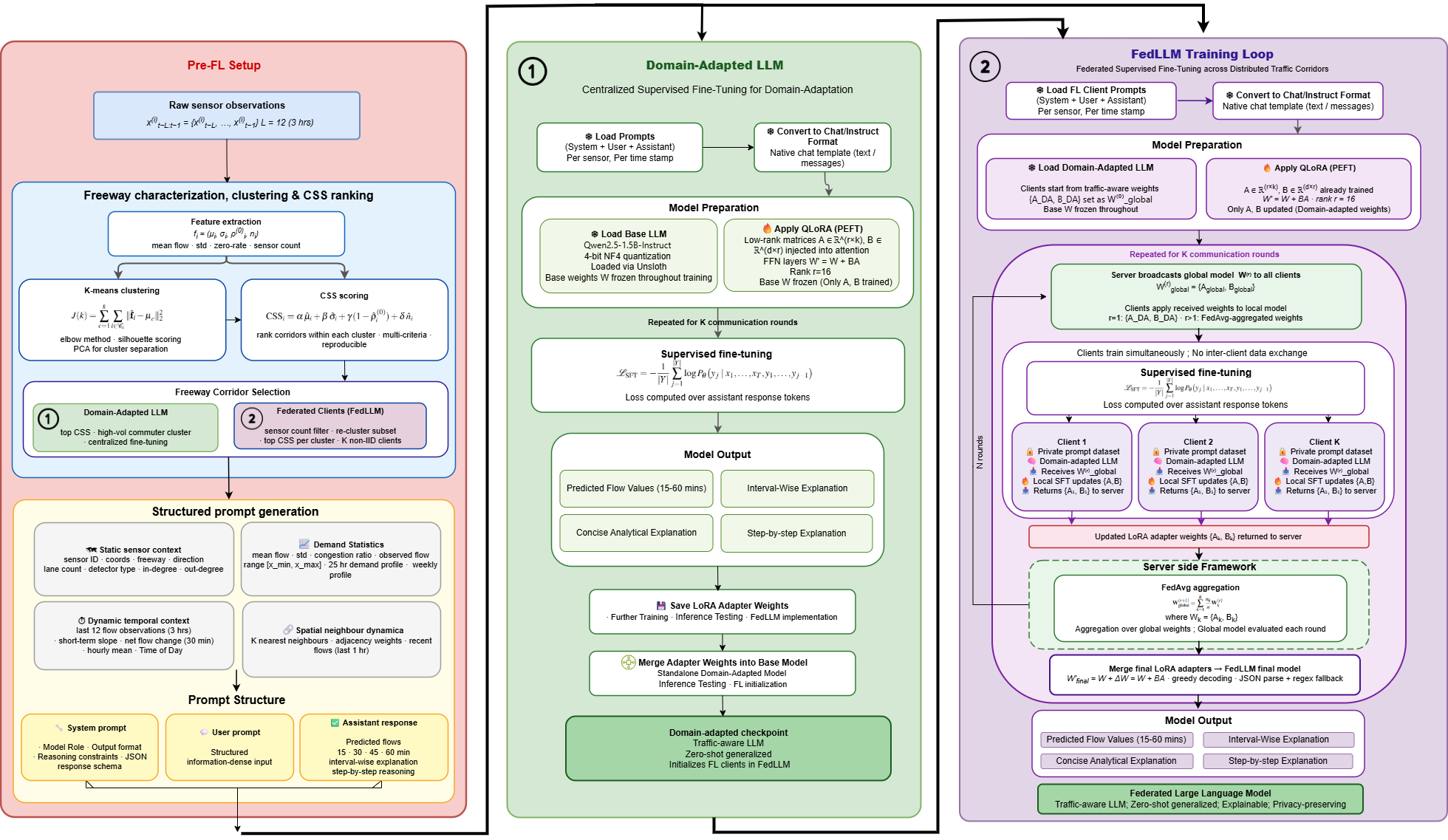}
    \caption[System architecture of the proposed FedLLM framework, illustrating the end-to-end pipeline]
    {\centering {System architecture of the proposed FedLLM framework, illustrating the end-to-end pipeline from data-driven corridor selection and structured prompt generation, through QLoRA-based domain adaptation, to privacy-preserving federated fine-tuning with explainable multi-horizon traffic flow prediction}}
    \label{fig:ArcChapter4}
\end{figure*}
\subsection{Federated LLM (FedLLM)}
\label{FL+LLM}
To address privacy and data-sharing limitations in traffic flow prediction, we extend the domain-adapted LLM with an FL framework that allows collaborative training across multiple freeway corridors without centralizing raw data. The proposed architecture follows a client-server setup consisting of a central coordinating server and four distributed clients, where each client represents a geographically distinct traffic environment and maintains its own locally generated traffic prompts. The federated simulation is implemented using Flower (flwr)~\cite{Beutel2020} framework, which provides flexible strategy implementation and a low-boilerplate setup that reduces communication overhead. It allows seamless orchestration of both server-side aggregation and client-side training across decentralized nodes and supports a wide range of ML libraries, including TensorFlow and PyTorch, making it accessible to developers. Each client initializes the model from the domain-adapted LoRA checkpoint obtained in the previous stage, so that federated training begins from a traffic-aware model rather than a general-purpose language model. This design choice helps local models converge faster and reduces the number of communication rounds required. During training, the base model weights remain frozen and only the LoRA adapter parameters $\Delta\mathbf{W} =\mathbf{B}\mathbf{A}$ are updated locally and exchanged with the server, reducing communication overhead while ensuring that raw traffic observations never leave the client.
For each configuration, client training data is randomly sampled using seed 3407 to ensure reproducibility across runs. At each communication round, the server broadcasts the current global LoRA weights to all clients. Each client applies these weights to its local model and performs fine-tuning on its private instructions-transformed traffic dataset for 200 optimization steps using the SFT objective defined in Equation~\ref{eq:sft_loss}. Local training uses an effective batch size of 16, a learning rate of $2 \times10^{-4}$ with cosine decay scheduling, the fused AdamW optimizer, and bfloat16 precision. It is worth noting that all clients perform local fine-tuning simultaneously within each communication round, making the training process fully distributed across independent nodes with no inter-client data exchange at any stage. After local optimization completes, the updated LoRA weights are returned to the server. The server then aggregates the client updates using the FedAvg algorithm \citep{McMahan2016}. Given $K = 4$ clients with local LoRA updates $\mathbf{W}_k$ and dataset sizes $n_k$, the global model for the next round is computed as 
\begin{equation} 
\mathbf{W}_{\text{global}}^{(r+1)} = 
\sum_{k=1}^{K} \frac{n_k}{n} \mathbf{W}_k^{(r)} 
\label{eq:fedavg} 
\end{equation} 
where $n = \sum_{k=1}^{K} n_k$ is the total number of training samples across all participating clients. Aggregation is applied exclusively to the LoRA adapter parameters while the base model weights remain fixed. The resulting global weights are then broadcast to all clients to begin the next round of training. This process is repeated for two communication rounds, after which the final LoRA weights are merged into the base model by adding the learned adapter updates $\Delta\mathbf{W}$ to the frozen base weights, producing the final federated traffic forecasting model for inference and evaluation. After each communication round, the global model is evaluated on the test dataset. Predictions are generated autoregressively using greedy decoding and extracted from the structured JSON outputs using a two-strategy extraction procedure that combines full JSON parsing with a regex fallback to handle truncated responses, ensuring robust prediction extraction across all clients. Performance is reported per client and per prediction horizon (15, 30, 45, and 60 minutes) using RMSE, MAE, MAPE, and R$^2$. Global federated metrics are obtained through a sample-weighted average across clients. The generated outputs demonstrate contextual awareness of freeway-specific demand patterns, while also accounting for nearby traffic conditions, congestion levels, and time-of-day variations. This indicates that the structured prompt design continues to perform consistently in the federated setting.

\textbf{Figure~\ref{fig:ArcChapter4}} further outlines the three-stage architecture of the proposed framework. Freeway corridors are first characterized through four structural descriptors computed from raw sensor observations, normalized, and grouped into distinct traffic regimes via K-means. CSS scoring then identifies the strongest candidate from each cluster, and the selected observations are converted into structured prompts that bring together sensor metadata, demand history, temporal context, and neighbouring flow activity. These prompts are used to fine-tune Qwen2.5-1.5B through QLoRA, where parameter updates are restricted to the injected low-rank matrices while the remainder of the model stays frozen, and the resulting checkpoint is passed to all four federated clients as their common starting point. Each client then works through its own private dataset independently, sending adapter updates back to the Flower server after each local training run. The server then aggregates these through FedAvg and redistributes the updated weights for the next round. Once training is done, the adapters are folded back into the base model, giving a forecasting model that pairs numeric flow predictions across four horizons with natural language reasoning.

\section{RESULTS AND DISCUSSION}
\label{Results_section}
This section presents a comprehensive evaluation of the two proposed models - the domain-adapted LLM and FedLLM framework for short-term traffic flow prediction across four horizons (15, 30, 45, and 60 minutes). Beyond predictive accuracy, this evaluation also examines scalability, cross-region generalization, privacy preservation, and 
explainability. Both models are compared against centralized and federated baselines, with the FedLLM further improving upon the domain-adapted variant. We also examine performance consistency across varying sample sizes and evaluate zero-shot cross-region transferability to assess robustness and practical applicability.
To measure forecasting performance, we adopt four standard metrics: Mean Absolute Error (MAE), Root Mean Squared Error (RMSE), Mean Absolute Percentage Error (MAPE), and the Coefficient of Determination (R$^2$). Let $y_i$ denote the ground truth and $\hat{y}_i$ the predicted value for the $i$-th sample, $n$ the total number of samples, and 
$\bar{y} = \frac{1}{n}\sum_{i=1}^{n} y_i$ the mean of the ground truth values.
\begin{equation}
\text{MAE} = \frac{1}{n} \sum_{i=1}^{n} 
\left| y_i - \hat{y}_i \right|
\label{eq:mae}
\end{equation}
\begin{equation}
\text{RMSE} = \sqrt{\frac{1}{n} \sum_{i=1}^{n} 
\left( y_i - \hat{y}_i \right)^2}
\label{eq:rmse}
\end{equation}
\begin{equation}
\text{MAPE} = \frac{1}{n} \sum_{i=1}^{n} 
\left| \frac{y_i - \hat{y}_i}{y_i} \right| \times 100\%
\label{eq:mape}
\end{equation}
\begin{equation}
R^2 = 1 - \frac{\sum_{i=1}^{n} (y_i - \hat{y}_i)^2}
{\sum_{i=1}^{n} (y_i - \bar{y})^2}
\label{eq:r2}
\end{equation}
where, MAE measures the average magnitude of prediction errors, RMSE penalizes larger deviations more strongly, MAPE expresses error as a percentage of actual flow, and $R^2$ quantifies how well the model explains the variability in the data. Using all four metrics provides a balanced evaluation of both accuracy and robustness under heterogeneous traffic conditions.
All experiments were conducted on a single NVIDIA A100-SXM4-80GB GPU using Google Colab, showing that the entire framework can be run with standard cloud-based resources without the need for specialized infrastructure. Each experiment was executed twice to check for consistency in the results. Full details of dataset partitioning, freeway selection, and train/test splits are provided in Section~\ref{data_preperation}. This setup reflects the practical feasibility of deploying the proposed framework under realistic computational constraints.

\subsection{Baseline Models}
We compare the proposed models against multiple baselines covering a range of temporal, spatio-temporal, and federated architectures.
Among the centralized baselines, GRU \citep{Cho2014} is an RNN used as a temporal-only recurrent model that captures sequential traffic patterns but does not account for spatial relationships. FC-LSTM \citep{Sutskever2014} 
similarly applies a fully connected LSTM to flow sequences without any spatial component. STGCN \citep{Yu2018} and DCRNN \citep{Li2018} represent graph-based spatio-temporal methods, where STGCN combines graph convolution with temporal convolution, while DCRNN models traffic propagation as a diffusion process on a directed road graph. AGCRN \citep{Bai2020} extends this further by learning node-specific graph structure adaptively rather 
than relying on a fixed adjacency matrix. ASTGNN\citep{Guo2022} adds dynamic spatial attention on top of temporal convolution to capture evolving inter-sensor dependencies. The centralized Qwen baseline \citep{Qwen2025} uses the same Qwen2.5-1.5B base model without any domain adaptation, directly comparing the contribution of fine-tuning on structured traffic prompts.
On the federated side, Fed-GDAN \citep{Li2025} is a graph diffusion attention network trained under federated aggregation, and FedASTGCN \citep{Zhang2021} applies a topology-aware FL approach for traffic forecasting. Both exchange full model weights at each communication round, whereas the proposed FedLLM (Section: \ref{FL+LLM}) exchanges only LoRA adapter parameters, thus reducing per-round communication cost considerably. All baselines follow configurations from their original publications, trained on the full available dataset and executed twice to confirm the consistency of results. 

\begin{table*}[htbp]
\centering
\scriptsize
\caption{Performance comparison of the domain-adapted LLM 
model against baseline methods across multiple prediction horizons. 
Best results are shown in \textbf{bold}.}
\label{tab:baseline_comparison}
\resizebox{\textwidth}{!}{%
\begin{tabular}{llcccccccc}
\toprule
\textbf{Time} & \textbf{Metric} & \textbf{GRU} & 
\textbf{STGCN} & \textbf{DCRNN} & \textbf{AGCRN} & 
\textbf{FC-LSTM} & \textbf{ASTGNN} & 
\textbf{Centralized Qwen} & 
\textbf{Domain-Adapted LLM} \\
\midrule
\multirow{4}{*}{15 min}
 & RMSE & 42.14 & 35.20 & 58.02 & 113.44 & 38.48 & 43.87 & 29.06 & \textbf{24.76} \\
 & MAE  & 27.59 & 26.51 & 42.99 & 75.85  & 25.82 & 32.95 & 20.18 & \textbf{17.08} \\
 & R$^2$ & 0.910 & 0.930 & 0.893 & 0.574  & 0.950 & 0.937 & 0.905 & \textbf{0.981} \\
 & MAPE & 14.99\% & 18.59\% & 35.97\% & 25.58\% & 24.82\% & 22.31\% & 20.80\% & \textbf{10.57\%} \\
\midrule
\multirow{4}{*}{30 min}
 & RMSE & 43.18 & 41.00 & 58.57 & 113.07 & 42.55 & 53.64 & 35.95 & \textbf{32.16} \\
 & MAE  & 28.56 & 30.64 & 44.10 & 73.37  & 28.19 & 38.31 & 26.40 & \textbf{22.27} \\
 & R$^2$ & 0.913 & 0.920 & 0.891 & 0.577  & 0.940 & 0.905 & 0.854 & \textbf{0.968} \\
 & MAPE & 15.34\% & 20.42\% & 44.41\% & 22.66\% & 25.25\% & 24.58\% & 24.26\% & \textbf{14.41\%} \\
\midrule
\multirow{4}{*}{45 min}
 & RMSE & 44.20 & 45.67 & 65.89 & 113.87 & 46.54 & 64.03 & 47.06 & \textbf{37.89} \\
 & MAE  & 29.29 & 34.06 & 50.21 & 74.06  & 30.57 & 44.01 & 35.29 & \textbf{26.75} \\
 & R$^2$ & 0.910 & 0.900 & 0.863 & 0.571  & 0.930 & 0.865 & 0.750 & \textbf{0.955} \\
 & MAPE & \textbf{15.73\%} & 22.30\% & 48.73\% & 22.92\% & 25.91\% & 26.96\% & 30.27\% & 18.26\% \\
\midrule
\multirow{4}{*}{60 min}
 & RMSE & 45.90 & 50.39 & 75.21 & 115.79 & 51.13 & 75.39 & 60.14 & \textbf{44.76} \\
 & MAE  & \textbf{30.61} & 37.08 & 57.50 & 77.95  & 33.28 & 50.43 & 46.21 & 31.41 \\
 & R$^2$ & 0.904 & 0.880 & 0.821 & 0.556  & 0.910 & 0.813 & 0.590 & \textbf{0.937} \\
 & MAPE & \textbf{16.65\%} & 22.94\% & 52.33\% & 26.08\% & 25.83\% & 29.28\% & 37.74\% & 20.82\% \\
\midrule
\multirow{4}{*}{Overall}
 & RMSE & 43.88 & 43.43 & 64.80 & 114.04 & 44.92 & 60.38 & 44.63 & \textbf{35.66} \\
 & MAE  & 29.01 & 32.07 & 48.70 & 75.31  & 29.47 & 41.42 & 32.02 & \textbf{24.38} \\
 & R$^2$ & 0.911 & 0.910 & 0.867 & 0.569  & 0.930 & 0.880 & 0.775 & \textbf{0.960} \\
 & MAPE & \textbf{15.68\%} & 21.06\% & 45.36\% & 24.31\% & 25.45\% & 25.78\% & 28.26\% & 16.02\% \\
\bottomrule
\end{tabular}%
}
\end{table*}

\subsection{Domain-Adapted LLM}
Table~\ref{tab:baseline_comparison} presents the performance of the domain-adapted LLM against six DL baselines and a centralized Qwen variant across all prediction horizons. The task is to utilize the past three hours of traffic data (12 time steps at 15-minute intervals) to forecast flow at 15, 30, 45, and 60 minutes ahead. All DL baselines are trained on the full dataset, while the domain-adapted LLM and centralized Qwen are evaluated on a sample of prompts, with 5,000 representing the largest configuration tested. This is sufficient for LLMs because their pre-trained general knowledge from large corpora reduces dependence on exhaustive local training data, unlike graph and recurrent models that must learn spatial and temporal structure entirely from scratch. The results further show that the domain-adapted Qwen model consistently outperforms or remains competitive with all baselines, underscoring the effectiveness of domain-specific fine-tuning.

As evident in Table~\ref{tab:baseline_comparison}, our model achieves the best overall RMSE of 35.66, MAE of 24.38, and R$^2$ of 0.960, indicating improved accuracy and a stronger ability to capture variability in traffic patterns compared to baseline methods. 
Among the centralized DL baselines, STGCN records the second-lowest overall RMSE of 43.43. The domain-adapted LLM improves upon this by reducing RMSE by 17.9\% relative to STGCN. This advantage is most pronounced in MAE, MAPE, and RMSE, where the margins are largest. Rather than relying on architectural complexity or additional training volume, this improvement comes from richer input prompts. While the baselines operate on raw numerical sequences and infer context implicitly, the domain-adapted LLM receives structured input encoding sensor metadata, long-run demand statistics, spatial neighbour dynamics, and time-of-day semantics, enabling it to reason about the conditions underlying traffic behaviour at a specific sensor and time.
At 15 minutes, it can further be observed that the domain-adapted LLM achieves RMSE of 24.76 and MAPE of 10.57\%, representing improvements of 14.9\% in RMSE over the next best result of 29.06 (centralized Qwen) and 29.5\% in MAPE over the next best of 14.99\% (GRU). The R$^2$ of 0.981 at this horizon indicates the model accounts for over 98\% of short-term flow variance. The proposed model's R$^2$ declines by only 0.044 from 15 to 60 minutes (0.981 $\to$ 0.937), compared to a decline of 0.124 for ASTGNN and 0.315 for the centralized Qwen, the steepest degradation of any model in the comparison. This indicates that domain adaptation improves long-horizon stability, with the statistical summaries embedded in each prompt continuing to anchor predictions to the sensor's established behaviour as the forecast window widens.
It is also worth noting that GRU records a marginally lower 
overall MAPE of 15.68\% compared to 16.02\% for the 
domain-adapted LLM, and a lower MAE of 30.61 against 
31.41 at the 60-minute horizon. These differences are 
attributable to MAPE's and MAE's sensitivity to 
low-flow overnight periods, where small absolute 
errors in near-zero flow conditions disproportionately 
influence percentage and absolute error metrics. Across 
RMSE and R$^2$, which provide a more complete picture of accuracy, the domain-adapted LLM maintains a consistent advantage 
at every horizon. Similarly, DCRNN records MAPE of 35.97\% at 15 minutes rising to 52.33\% at 60 minutes, reflecting sensitivity 
to low-flow conditions that the explicit time-of-day 
and congestion encoding in the domain-adapted LLM's prompt helps to mitigate. MAE follows a consistent progression from 17.08 at 
15 minutes to 31.41 at 60 minutes, remaining below 
every baseline at each horizon with the exception of 
GRU at 60 minutes, as discussed above.

\begin{table*}[htbp]
\centering
\caption{Performance of the domain-adapted Qwen model across 
different evaluation sample sizes, assessing the consistency of 
results.}
\label{tab:test_size}
\begin{tabular}{llcccc}
\toprule
\textbf{Time Interval} & \textbf{Metric} & 
\textbf{100 Samples} & \textbf{500 Samples} & 
\textbf{1000 Samples} & \textbf{5000 Samples} \\
\midrule
\multirow{4}{*}{15 min}
 & RMSE & 24.46 & 24.90 & 24.70 & 24.76 \\
 & MAE  & 16.36 & 16.95 & 16.79 & 17.08 \\
 & R$^2$ & 0.980 & 0.980 & 0.980 & 0.981 \\
 & MAPE & 8.29\% & 12.35\% & 10.64\% & 10.57\% \\
\midrule
\multirow{4}{*}{30 min}
 & RMSE & 37.71 & 32.67 & 31.71 & 32.16 \\
 & MAE  & 22.24 & 22.57 & 21.91 & 22.27 \\
 & R$^2$ & 0.960 & 0.960 & 0.960 & 0.968 \\
 & MAPE & 14.79\% & 17.97\% & 14.54\% & 14.41\% \\
\midrule
\multirow{4}{*}{45 min}
 & RMSE & 39.19 & 35.89 & 36.30 & 37.89 \\
 & MAE  & 26.85 & 25.60 & 26.10 & 26.75 \\
 & R$^2$ & 0.950 & 0.960 & 0.950 & 0.955 \\
 & MAPE & 16.81\% & 16.42\% & 16.08\% & 18.26\% \\
\midrule
\multirow{4}{*}{60 min}
 & RMSE & 43.54 & 42.85 & 43.53 & 44.76 \\
 & MAE  & 31.23 & 30.14 & 30.96 & 31.41 \\
 & R$^2$ & 0.940 & 0.940 & 0.940 & 0.937 \\
 & MAPE & 20.65\% & 23.88\% & 21.72\% & 20.82\% \\
\midrule
\multirow{4}{*}{Overall}
 & RMSE & 36.92 & 34.68 & 34.14 & 35.66 \\
 & MAE  & 24.17 & 23.82 & 23.94 & 24.38 \\
 & R$^2$ & 0.960 & 0.960 & 0.960 & 0.960 \\
 & MAPE & 15.14\% & 17.65\% & 15.75\% & 16.02\% \\
\bottomrule
\end{tabular}
\end{table*}
\begin{table*}[htbp]
\centering
\caption{Zero-shot cross-region generalization performance of our domain-adapted LLM evaluated on GBA District 4 
across different test sample sizes.}
\label{tab:zero_shot}
\begin{tabular}{llccc}
\toprule
\textbf{Time Interval} & \textbf{Metric} & 
\textbf{1000 Samples} & \textbf{2000 Samples} & 
\textbf{5000 Samples} \\
\midrule
\multirow{4}{*}{15 min}
 & RMSE & 28.61 & 30.07 & 28.70 \\
 & MAE  & 18.58 & 19.29 & 18.70 \\
 & R$^2$ & 0.960 & 0.962 & 0.966 \\
 & MAPE & 11.24\% & 12.17\% & 12.38\% \\
\midrule
\multirow{4}{*}{30 min}
 & RMSE & 35.17 & 36.03 & 34.84 \\
 & MAE  & 23.45 & 24.35 & 23.43 \\
 & R$^2$ & 0.940 & 0.947 & 0.951 \\
 & MAPE & 13.83\% & 15.38\% & 16.18\% \\
\midrule
\multirow{4}{*}{45 min}
 & RMSE & 41.73 & 41.63 & 40.99 \\
 & MAE  & 28.56 & 28.73 & 28.09 \\
 & R$^2$ & 0.920 & 0.928 & 0.931 \\
 & MAPE & 17.76\% & 17.89\% & 19.26\% \\
\midrule
\multirow{4}{*}{60 min}
 & RMSE & 47.64 & 48.26 & 46.94 \\
 & MAE  & 33.47 & 34.01 & 32.66 \\
 & R$^2$ & 0.900 & 0.904 & 0.909 \\
 & MAPE & 20.72\% & 21.31\% & 22.61\% \\
\midrule
\multirow{4}{*}{Overall}
 & RMSE & 38.94 & 39.57 & 38.47 \\
 & MAE  & 26.01 & 26.59 & 25.72 \\
 & R$^2$ & 0.930 & 0.935 & 0.939 \\
 & MAPE & 15.89\% & 16.69\% & 17.61\% \\
\bottomrule
\end{tabular}
\end{table*}
The value of domain adaptation is most directly illustrated against the centralized Qwen baseline, which shares the same 1.5B parameter base model, prompt format, and inference procedure, with fine-tuning being the only distinction. As evident from the results, the domain-adapted model consistently achieves stronger results across all prediction horizons. This indicates that while pre-trained language models provide a useful starting point, domain-specific adaptation is necessary to align predictions with the underlying traffic dynamics and produce physically meaningful forecasts.
\begin{table*}[htbp]
\centering
\caption{Performance comparison of FedLLM against federated 
and centralized baselines across four prediction horizons. 
Baseline models follow standard configurations reported in 
original publications. Best results are shown in \textbf{bold}.}
\label{tab:fl_comparison}
\resizebox{\textwidth}{!}{%
\begin{tabular}{llcccccccccc}
\toprule
\textbf{Time} & \textbf{Metric} & 
\textbf{GRU} & \textbf{STGCN} & \textbf{DCRNN} & 
\textbf{ASTGNN} & \textbf{FC-LSTM} & 
\textbf{Fed-GDAN} & \textbf{FedASTGCN} & 
\textbf{Centralized Qwen} & 
\textbf{Domain-Adapted Qwen} & 
\textbf{FedLLM} \\
\midrule
\multirow{4}{*}{15 min}
 & RMSE  & 37.79 & 42.73 & 48.73 & 31.82 & 30.32 & 42.16 & 25.13 & 29.06 & 19.03 & \textbf{16.30} \\
 & MAE   & 22.66 & 34.22 & 33.01 & 21.12 & 17.72 & 27.55 & 17.72 & 20.18 & 12.00 & \textbf{10.56} \\
 & R$^2$ & 0.860 & 0.939 & 0.932 & 0.971 & 0.973 & 0.936 & 0.946 & 0.905 & 0.986 & \textbf{0.989} \\
 & MAPE  & 22.67\% & 24.01\% & 94.79\% & 25.42\% & 28.97\% & 66.02\% & 32.82\% & 20.80\% & 18.25\% & \textbf{16.44\%} \\
\midrule
\multirow{4}{*}{30 min}
 & RMSE  & 40.46 & 47.19 & 52.71 & 40.08 & 33.90 & 44.20 & 33.72 & 35.95 & 26.63 & \textbf{21.33} \\
 & MAE   & 24.27 & 36.55 & 35.95 & 25.63 & 19.82 & 28.98 & 23.20 & 26.40 & 15.86 & \textbf{13.80} \\
 & R$^2$ & 0.850 & 0.917 & 0.920 & 0.954 & 0.967 & 0.929 & 0.898 & 0.854 & 0.977 & \textbf{0.985} \\
 & MAPE  & 24.00\% & 24.84\% & 88.60\% & 28.72\% & 32.28\% & 67.17\% & 36.22\% & 24.26\% & 22.84\% & \textbf{19.27\%} \\
\midrule
\multirow{4}{*}{45 min}
 & RMSE  & 43.11 & 50.40 & 59.96 & 48.86 & 37.73 & 47.06 & 42.52 & 47.06 & 32.74 & \textbf{25.42} \\
 & MAE   & 26.37 & 40.92 & 40.78 & 30.81 & 21.90 & 31.24 & 29.00 & 35.29 & 19.27 & \textbf{16.61} \\
 & R$^2$ & 0.830 & 0.932 & 0.897 & 0.932 & 0.959 & 0.923 & 0.833 & 0.750 & 0.967 & \textbf{0.983} \\
 & MAPE  & 26.17\% & 27.02\% & 91.90\% & 33.12\% & 36.94\% & 69.04\% & 40.54\% & 30.27\% & 25.63\% & \textbf{24.27\%} \\
\midrule
\multirow{4}{*}{60 min}
 & RMSE  & 46.88 & 47.55 & 70.87 & 58.31 & 42.25 & 50.74 & 51.21 & 60.14 & 37.88 & \textbf{30.18} \\
 & MAE   & 29.08 & 38.20 & 49.13 & 36.48 & 24.27 & 33.90 & 34.69 & 46.21 & 22.74 & \textbf{19.70} \\
 & R$^2$ & 0.810 & 0.935 & 0.856 & 0.903 & 0.948 & 0.907 & 0.753 & 0.590 & 0.956 & \textbf{0.983} \\
 & MAPE  & 28.58\% & \textbf{23.75\%} & 105.98\% & 38.10\% & 41.15\% & 71.45\% & 45.72\% & 37.74\% & 31.24\% & 27.37\% \\
\midrule
\multirow{4}{*}{Overall}
 & RMSE  & 42.14 & 46.97 & 58.07 & 45.84 & 36.32 & 46.15 & 42.91 & 44.63 & 29.91 & \textbf{23.31} \\
 & MAE   & 25.60 & 37.47 & 39.72 & 28.51 & 20.93 & 30.44 & 26.57 & 32.02 & 17.47 & \textbf{15.07} \\
 & R$^2$ & 0.843 & 0.930 & 0.901 & 0.940 & 0.962 & 0.923 & 0.947 & 0.775 & 0.974 & \textbf{0.985} \\
 & MAPE  & 25.36\% & 24.91\% & 95.22\% & 31.34\% & 34.83\% & 68.42\% & 37.83\% & 28.26\% & 24.52\% & \textbf{21.84\%} \\
\bottomrule
\end{tabular}%
}
\end{table*}
Table~\ref{tab:test_size} further evaluates the model across multiple test sets ranging from 100 to 5,000 samples. At 15 minutes, RMSE varies between 24.46 and 24.90, despite a substantial increase in sample size. While performance does not increase monotonically with larger test sets, this reflects genuine evaluation stability as the model produces consistent results. This is 
particularly relevant for real-world deployments where data availability can vary significantly across regions. The domain-adapted LLM can therefore be effectively evaluated with only a few samples, making it well-suited for new freeways where only limited historical data may be available.

Table~\ref{tab:zero_shot} reports performance on GBA District 4 under zero-shot conditions, with no data from this region included at any stage of training. We evaluate across three test sizes (1,000, 2,000, and 5,000 samples) to ensure that observed performance is not dependent on a particular sample selection. The region differs from the training data in freeway structure, sensor density, demand patterns and several other characteristics. Despite these differences, the model achieves overall R$^2$ of 0.930--0.939, with R$^2$ of 0.960--0.966 at 15 minutes and 0.900--0.909 at 60 minutes, suggesting that the model captures general traffic patterns that extend beyond the training region. The use of features such as mean demand, congestion levels, temporal profiles, and spatial relationships provides a consistent representation of traffic behaviour, allowing the model to be applied to new corridors 
without additional training.
Furthermore, as shown in Figures~\ref{fig:prompt_example_1} and \ref{fig:prompt_example_2}, each prediction is accompanied by a structured explanation, including interval-wise flow estimates across all prediction horizons, a concise summary of current traffic conditions, and a step-by-step description linking the prediction to observed temporal trends, congestion levels, and neighbouring sensor behaviour. This inclusion allows the predictions to be examined in the context of the input data, which is useful when interpreting model behaviour in practical traffic analysis settings.
To summarise, the domain-adapted LLM demonstrates strong and stable performance for multi-horizon traffic forecasting. The model maintains consistent accuracy even when evaluated on smaller sample sizes, indicating robustness under limited data availability. It also generalizes effectively to an unseen geographic region without requiring additional training. In addition to predictive performance, the model provides structured explanations alongside each forecast, grounded in the observed sensor context. This combination of accuracy, generalization, and interpretability highlights the potential of domain-adapted LLMs as a practical framework for traffic flow prediction.

\begin{table*}[t]
\centering
\small
\caption{Ablation study examining the effect of training data scale on the performance of the proposed FedLLM framework across multiple forecasting horizons.}
\begin{tabular}{llccc}
\toprule
\textbf{Time Interval} & \textbf{Metric} & 
\begin{tabular}[c]{@{}c@{}}Train/Test (1000/500)\end{tabular} &
\begin{tabular}[c]{@{}c@{}}Train/Test (2000/1000)\end{tabular} &
\begin{tabular}[c]{@{}c@{}}Train/Test (5000/3000)\end{tabular} \\
\midrule

\multirow{4}{*}{15 min} 
& RMSE & 16.30 & 16.54 & 16.04 \\
& MAE  & 10.56 & 10.79 & 10.40 \\
& R$^2$ & 0.989 & 0.963 & 0.972 \\
& MAPE & 16.44\% & 17.95\% & 17.04\% \\
\midrule

\multirow{4}{*}{30 min} 
& RMSE & 21.33 & 22.01 & 22.23 \\
& MAE  & 13.80 & 14.13 & 14.23 \\
& R$^2$ & 0.985 & 0.951 & 0.948 \\
& MAPE & 19.27\% & 21.30\% & 20.40\% \\
\midrule

\multirow{4}{*}{45 min} 
& RMSE & 25.42 & 27.62 & 27.23 \\
& MAE  & 16.61 & 17.08 & 17.29 \\
& R$^2$ & 0.983 & 0.922 & 0.922 \\
& MAPE & 24.27\% & 24.82\% & 24.30\% \\
\midrule

\multirow{4}{*}{60 min} 
& RMSE & 30.18 & 31.46 & 31.30 \\
& MAE  & 19.70 & 19.85 & 20.11 \\
& R$^2$ & 0.983 & 0.900 & 0.898 \\
& MAPE & 27.37\% & 28.29\% & 27.32\% \\
\midrule

\multirow{4}{*}{Overall} 
& RMSE & 23.31 & 24.40 & 24.20 \\
& MAE  & 15.07 & 15.46 & 15.51 \\
& R$^2$ & 0.985 & 0.936 & 0.935 \\
& MAPE & 21.84\% & 23.09\% & 22.26\% \\
\bottomrule
\end{tabular}
\label{tab:fl_llm_scaling}
\end{table*}

\subsection{FedLLM}
Table~\ref{tab:fl_comparison} evaluates the FedLLM framework against six centralized DL baselines, two federated baselines, the centralized Qwen, and the domain-adapted LLM across all prediction horizons. Consistent with the previous section, all DL and federated baselines are trained and evaluated on the full dataset. Meanwhile, the centralized Qwen, domain-adapted LLM, and FedLLM are evaluated on the same sampled prompt configurations, ensuring a fair comparison across the three LLM-based approaches. The FedLLM results reported in Table~\ref{tab:fl_comparison} correspond to the best-performing configuration identified through the ablation study in Table~\ref{tab:fl_llm_scaling}. 
This is sufficient for two reasons. First, LLMs carry broad linguistic and reasoning knowledge from large-scale pre-training, reducing their dependence on local data volume compared to models that learn from scratch. Second, each client is initialized from the domain-adapted LLM checkpoint, which already encodes traffic-specific patterns from domain-specific fine-tuning, meaning only light local adaptation is needed per client.
As described in Section~\ref{FL+LLM}, the federated training process and communication protocol are designed to keep raw traffic data local to each client at all times. The four clients reflect a non-IID data distribution across structurally diverse traffic environments, a common and challenging condition in FL that the proposed framework effectively addresses. To the best of our knowledge, this is the first implementation of a federated LLM framework for freeway traffic flow prediction, thus establishing a new direction for privacy-preserving, explainable, and distributed traffic forecasting.
\begin{figure*}[htbp]
    \centering
    \includegraphics[width=0.85\textwidth]{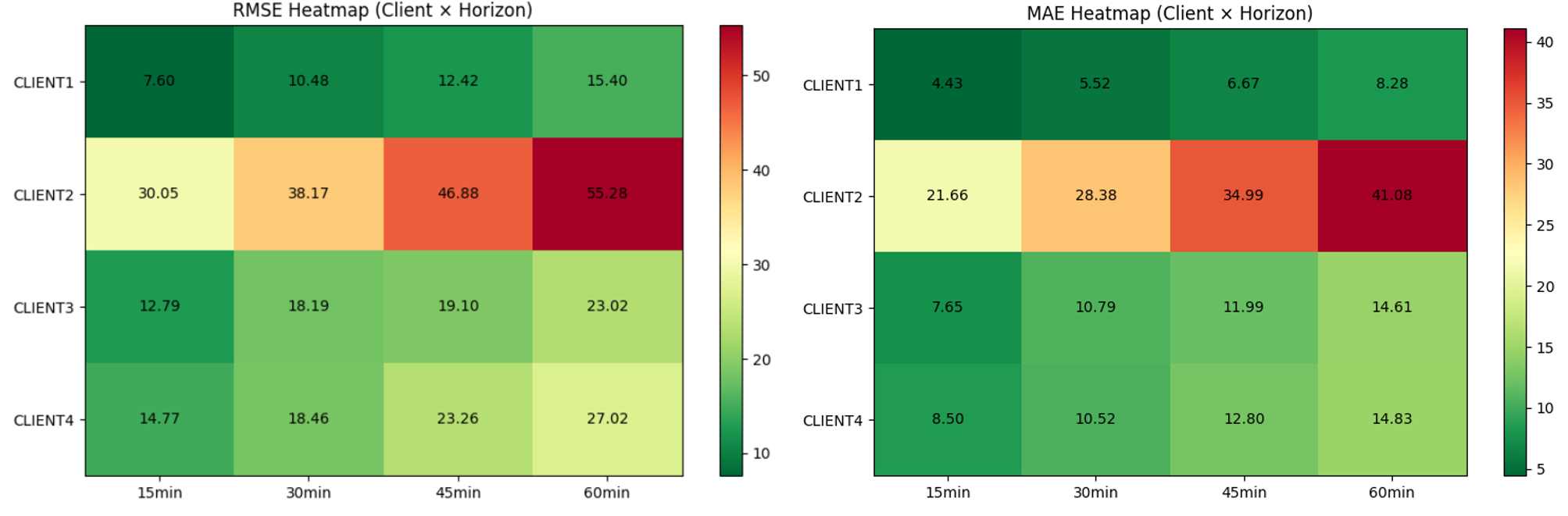}
    \caption[RMSE and MAE heatmaps of the FedLLM framework for different clients across four prediction horizons (15-60 min), ]{RMSE and MAE heatmaps of the FedLLM framework for different clients across four prediction horizons (15-60 min), illustrating consistent prediction fidelity despite 
    structurally heterogeneous client traffic conditions.}
    \label{fig:fl_heatmap}
\end{figure*}

\begin{figure*}[htbp]
    \centering
    \includegraphics[width=0.8\textwidth]{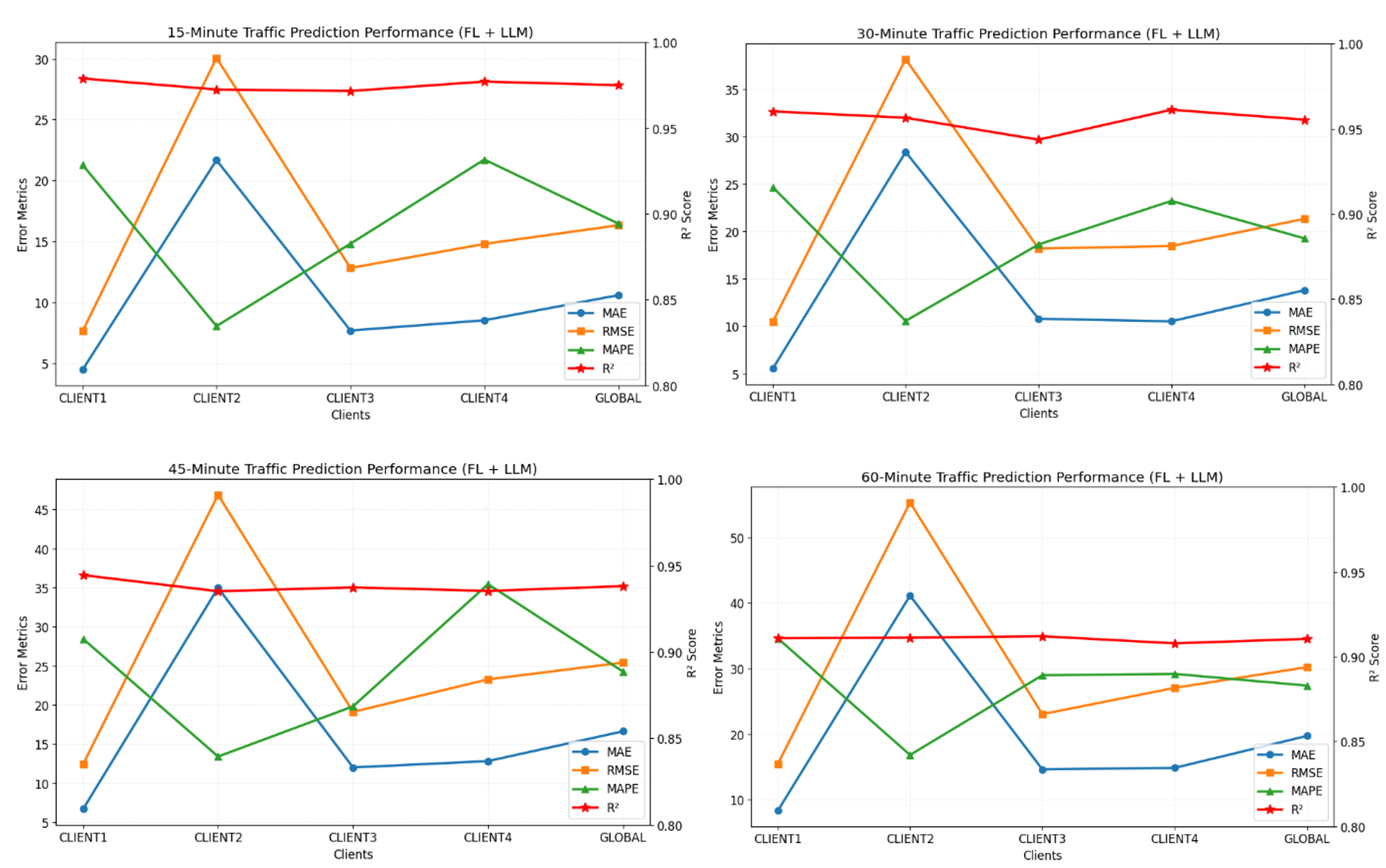}
    \caption[Client-wise and global evaluation metrics of the proposed FedLLM framework across prediction horizons (15–60 min),]{Client-wise and global evaluation metrics of the 
    proposed FedLLM framework across prediction horizons (15--60 min), illustrating stable global generalization under varying client traffic conditions.}
    \label{fig:fl_error_metrics}
\end{figure*}
As evident in Table~\ref{tab:fl_comparison}, the FedLLM achieves the best overall performance, with RMSE of 23.31, MAE of 15.07, and R$^2$ of 0.985, outperforming all centralized and federated baselines. 
Among the federated baselines, FedASTGCN provides the closest comparison (RMSE 42.91, R$^2$ 0.947), though the gap remains substantial. Meanwhile, compared to the domain-adapted LLM, FedLLM further reduces RMSE from 35.66 to 23.31 and improves R$^2$ from 0.960 to 0.985. The improvement is consistent across MAE and RMSE, indicating a clear gain in prediction accuracy. This behaviour can be attributed to the diversity of client-specific traffic conditions. Each client adapts the shared model to its local patterns, and the aggregation process combines these updates into a global model that reflects a broader range of traffic behaviours than a single centralized dataset. At the 15-minute horizon, the FedLLM achieves an RMSE of 16.30 and an R$^2$ of 0.989, corresponding to a 14.4\% reduction in RMSE compared to the next best result of 19.03 (domain-adapted Qwen). As the forecast horizon increases, performance remains stable. The R$^2$ decreases only slightly from 0.989 at 15 minutes to 0.983 at 60 minutes, representing the smallest degradation among all models considered. At 60 minutes, the model still achieves RMSE of 30.18 and MAE of 19.70, both lower than the best results reported by the baselines. These results suggest that the combination of domain-adapted initialization and federated fine-tuning helps maintain prediction quality across both short and long horizons. However, a different pattern is observed for DCRNN and Fed-GDAN, which exhibit notably high MAPE values. For instance, DCRNN reaches 94.79\% at 15 minutes and 105.98\% at 60 minutes, while Fed-GDAN exceeds 66\% across all horizons. These large values are likely influenced by low-flow periods, where small absolute errors lead to inflated percentage errors. In contrast, the FedLLM shows more stable behaviour across RMSE and R$^2$ at all horizons. Figure~\ref{fig:prompt_example_2} further shows that the FedLLM produces structured outputs that link predicted flows directly to the observed sensor context through a step-by-step reasoning chain.
\begin{table*}[htbp]
\centering
\caption{Zero-shot cross-region generalization performance of the FedLLM framework evaluated on GBA District 4 across 
different test sample sizes.}
\label{tab:fl_zero_shot}
\begin{tabular}{llccc}
\toprule
\textbf{Time Interval} & \textbf{Metric} & 
\textbf{1000 Samples} & \textbf{2000 Samples} & 
\textbf{5000 Samples} \\
\midrule
\multirow{4}{*}{15 min}
 & RMSE & 27.96 & 27.33 & 28.30 \\
 & MAE  & 18.69 & 18.47 & 18.56 \\
 & R$^2$ & 0.967 & 0.969 & 0.966 \\
 & MAPE & 11.34\% & 11.50\% & 12.48\% \\
\midrule
\multirow{4}{*}{30 min}
 & RMSE & 38.47 & 35.69 & 35.96 \\
 & MAE  & 25.24 & 24.11 & 24.12 \\
 & R$^2$ & 0.936 & 0.946 & 0.945 \\
 & MAPE & 16.19\% & 15.83\% & 16.41\% \\
\midrule
\multirow{4}{*}{45 min}
 & RMSE & 49.26 & 45.76 & 45.11 \\
 & MAE  & 31.73 & 30.26 & 30.12 \\
 & R$^2$ & 0.896 & 0.911 & 0.914 \\
 & MAPE & 19.11\% & 18.90\% & 21.39\% \\
\midrule
\multirow{4}{*}{60 min}
 & RMSE & 56.58 & 53.52 & 53.89 \\
 & MAE  & 37.53 & 36.09 & 36.28 \\
 & R$^2$ & 0.864 & 0.881 & 0.878 \\
 & MAPE & 22.57\% & 24.06\% & 24.84\% \\
\midrule
\multirow{4}{*}{Overall}
 & RMSE & 43.07 & 41.77 & 41.93 \\
 & MAE  & 28.25 & 27.23 & 27.27 \\
 & R$^2$ & 0.916 & 0.927 & 0.926 \\
 & MAPE & 17.30\% & 17.58\% & 18.78\% \\
\bottomrule
\end{tabular}
\end{table*}
The client-level analysis provides insight into how the FedLLM behaves under heterogeneous traffic conditions. The four clients correspond to structurally distinct corridors, as described in Section~\ref{data_preperation}, ranging from low-volume peripheral routes to high-flow commuter expressways, and together form a non-IID distribution representative of real-world deployments. As shown in Figure~\ref{fig:fl_heatmap}, prediction errors differ across clients but follow a consistent pattern across all horizons. The low-volume client exhibits the smallest absolute errors (RMSE of 7.60 and MAE of 4.43 at 15 minutes), whereas the high-flow client records larger values (RMSE 30.05 and MAE 21.66). This difference is expected, as higher traffic volumes lead to larger absolute deviations. When interpreted relative to the scale of each corridor, the model maintains consistent predictive performance across all clients.

Figure~\ref{fig:fl_error_metrics} further shows how MAE, RMSE, MAPE, and R$^2$ evolve with increasing prediction horizon. Across all clients, error metrics increase gradually from 15 to 60 minutes, while R$^2$ declines in a smooth and consistent manner. The global metrics closely track the average behaviour of individual clients, suggesting that the aggregated model does not bias towards any particular traffic regime. This consistency across a non-IID client distribution indicates that the federated model remains stable under heterogeneous data conditions, which is important for multi-agency scenarios where traffic characteristics vary across regions.
The effect of training and evaluation scale is examined in Table~\ref{tab:fl_llm_scaling}, which considers three configurations (train/test): 1,000/500, 2,000/1,000, and 5,000/3,000 samples per client. Evaluating both training and test sizes provides a more complete view of scalability. The 1,000/500 configuration achieves the best overall performance, with RMSE of 23.31 and R$^2$ of 0.985. Increasing the dataset size leads to only minor changes, with RMSE remaining within a narrow range and R$^2$ consistently above 0.935. Notably, even the smallest configuration outperforms all baselines in Table~\ref{tab:fl_comparison}, including FC-LSTM — the best performing non-LLM centralized baseline with an overall RMSE of 36.32. These results further indicate that our model maintains stable predictive accuracy as the dataset size increases, suggesting that the framework can achieve strong forecasting performance even with relatively limited training data. This highlights the data efficiency and robustness of our model for multi-horizon traffic prediction. With greater computational resources, future work will extend this to larger client populations, additional communication rounds, and increased dataset sizes to further improve global performance and cross-region transferability.

Table~\ref{tab:fl_zero_shot} further evaluates the ability of the model to generalize beyond the training distribution through zero-shot testing on GBA District 4, a region not included at any stage of training. As shown in the table, the performance is reported across three test sizes (1,000, 2,000, and 5,000 samples) to ensure that the results are not dependent on a specific sample subset. This region differs from the training corridors in terms of freeway structure, sensor density, and demand patterns. Despite these differences, the model achieves overall R$^2$ values between 0.916 and 0.927, with R$^2$ ranging from 0.966 to 0.969 at 15 minutes and from 0.864 to 0.881 at 60 minutes. Moreover, at the 15-minute horizon, the zero-shot result R$^2$ = 0.969 is only 0.020 lower than the model's 
performance on the original test set, indicating that the learned representations extend effectively to previously unseen traffic conditions. For context, this value exceeds the domain-adapted LLM's performance at 60 minutes under standard evaluation (R$^2$ = 0.937). A small decrease compared to the centralized domain-adapted model is expected, as federated training encourages corridor-specific adaptation. Even so, the model retains strong predictive performance without requiring retraining or access to data from the target region.

A central feature of the FedLLM framework is that raw traffic observations remain local to each client, as described in Section~\ref{FL+LLM}. This makes our model suitable for deployment scenarios where data sharing across agencies is restricted by governance or privacy constraints. Clients exchange only LoRA adapter parameters (approx 70.5 MB per round), rather than full model weights (2.9 GB in our case). This makes the FedLLM considerably more suitable for deployments with constrained network bandwidth or strict data governance policies.
Overall, the FedLLM framework demonstrates strong and stable performance across all evaluation settings, achieving the best results among all models compared with an overall R$^2$ of 0.985 and the smallest R$^2$ degradation from 15 to 60 minutes. Operating in a fully decentralized manner, it preserves data privacy and reduces communication overhead while maintaining predictive accuracy. The model generalizes effectively to an unseen geographic region without retraining, and every 
prediction is accompanied by a structured natural language explanation that no federated baseline provides. The complete pipeline runs on a single GPU via standard cloud platforms, demonstrating practical accessibility beyond institutional research settings.

\subsection{Domain-Adapted LLM vs FedLLM: Comparative Discussion}
Taken together, the results of the two proposed models show a clear progression. The domain-adapted LLM demonstrates that a compact instruction-tuned model, fine-tuned on structured traffic prompts, outperforms dedicated graph-based spatio-temporal architectures across most metrics and prediction horizons. The FedLLM builds on this further, achieving stronger results without centralized data access. This behaviour can be attributed to the diversity of traffic conditions observed across clients, where each client adapts the shared model to its local patterns, and aggregation combines these updates into a more representative global model. The two approaches are suited for different deployment settings. The domain-adapted LLM requires data collected from multiple corridors to be pooled into a single training set - practical within a single agency, but often not possible when data is held by separate authorities. The FedLLM addresses this limitation by enabling local training at each client, with only LoRA adapter parameters shared during aggregation. As a result, raw traffic observations remain local at all times. The federated design also brings a practical scalability advantage. 
Adding a new corridor does not require rebuilding a centralized dataset or retraining the model from scratch. Instead, a new client can initialize from the current global model, perform local fine-tuning, and participate in subsequent communication rounds. 
Both models offer the same level of interpretability. Every prediction comes with a structured explanation grounded in the input context, covering temporal trends, congestion conditions, and neighbouring sensor behaviour.  This is particularly relevant in operational settings, where decision-makers rely on both predictions and supporting reasoning. The FedLLM maintains this capability in a decentralized setting, whereas existing federated baselines considered in this study provide only numerical outputs without interpretive support.
More broadly, the results show that LLMs can serve as a flexible forecasting backbone when paired with structured prompt design. The same model can be applied in both centralized and federated settings. The key factor is the prompt representation, which encodes traffic context in a consistent and transferable format. This includes sensor metadata, demand statistics, spatial relationships, and temporal features. The ability of our FedLLM model to generalize across training setups further distinguishes the proposed models from conventional approaches and highlights the potential of prompt-driven LLM frameworks for privacy-preserving, interpretable, and scalable traffic forecasting.
\section{LIMITATIONS AND FUTURE DIRECTIONS}
\label{limitation_FutureWork}
While the proposed framework achieves promising results in terms of accuracy, robustness, generalization and privacy-preservation, several aspects offer scope for continued refinement. The following points outline the current limitations and the corresponding directions for future work to strengthen the methodology and practical applicability of our work.

\begin{enumerate}
\item Traffic-Aware Aggregation: The current FedLLM uses standard FedAvg \citep{McMahan2016}, where client updates are weighted by local sample size alone, without accounting for differences in traffic dynamics across corridors. Recent work has shown that more flexible aggregation strategies better handle heterogeneous data distributions \citep{Feng2024, Tang2023, Zhang2024a, Lu2024}, yet none have been applied within an LLM-based federated framework. Our prior work \citep{Kaur2026} showed that dynamically weighting client updates based on traffic-specific metrics improves both convergence and predictive performance under non-IID conditions. Future work will build on this design by integrating traffic-aware aggregation into the FedLLM framework. Approaches such as FedProx \citep{Karimireddy2019} and adaptive federated optimizers \citep{Reddi2020} will further be explored to improve global model quality across structurally diverse clients.

\item Scale of Training and Model Capacity: The framework achieves strong and stable performance using a compact 1.5B parameter model \citep{Qwen2025} trained on sampled prompt subsets, indicating effective data utilization. Expanding the training setup to include larger and more diverse prompt collections, additional communication rounds, and a greater number of federated clients would allow the model to capture a wider range of traffic patterns and variability. In parallel, evaluating larger instruction-tuned models such as LLaMA-3 \citep{Grattafiori2024} and DeepSeek-V3 \citep{deepseek} will provide insight into whether increased model capacity leads to further gains in forecasting accuracy and reasoning quality across both centralized and federated settings.

\item Dataset Diversity and Traffic Regimes: The framework is currently evaluated on freeway corridors within the California PeMS network \citep{CaltransWebsite}, where traffic is measured by loop detectors on large highway networks. This is a well-controlled and widely used setting, but it does not reflect the full range of real-world traffic environments. Urban arterial roads, signalized intersections, and mixed urban-highway networks each present distinct demand patterns, sensor configurations, and congestion characteristics that may require adaptation of the prompt design. Future work will extend evaluation to benchmarks including METR-LA, PEMS-BAY, PEMS07, and PEMS08~\citep{Liu2023a}, as well as urban road network datasets~\citep{Sowinski2016}. Prediction horizons beyond 60 minutes will also be examined to assess long-range temporal stability across diverse traffic regimes.
\end{enumerate}
\section{CONCLUSION}
\label{Conclusion}
This study presented a novel prompt-driven framework for freeway traffic flow prediction that integrates domain-adapted LLMs with FL. The work was motivated by two practical constraints that continue to limit the deployment of data-driven traffic forecasting models in real-world settings. Traffic data is typically administered by separate regional authorities under distinct ownership and governance policies, making centralized aggregation across agency boundaries infeasible in most operational contexts. At the same time, existing spatio-temporal models produce only numerical outputs, lacking interpretable reasoning that traffic operators can assess and act upon, a limitation that diminishes their practical utility in time-sensitive management decisions.
The proposed framework addressed both constraints by designing two complementary models. The domain-adapted LLM demonstrates that a compact instruction-tuned model, trained on structured traffic prompts encoding sensor metadata, spatial relationships, demand statistics, and temporal context, can achieve strong performance without relying on specialized graph architectures or large training volumes. Further, the proposed CSS score provides a principled method for selecting representative freeway corridors based on multiple traffic characteristics.
Building on this, the FedLLM extends the approach to a decentralized setting in which clients train locally and exchange only lightweight adapter parameters, improving performance while preserving data locality.
Both models generalize effectively to unseen regions without retraining, indicating that the prompt representation captures transferable traffic patterns across varying corridor structures and demand conditions. In addition, each prediction is accompanied by a structured natural language explanation, providing interpretable insights into predicted flow patterns, congestion conditions, and spatial context. 
From a deployment perspective, the framework is suited to environments where traffic data is distributed across independent agencies and cannot be centrally aggregated. Transportation authorities, freeway management units, and traffic control centres can each contribute to collaborative model improvement without exposing their local data to any external party. The inclusion of structured explanations alongside predictions supports more informed decision-making for routing, signal control, incident response, and infrastructure planning. The framework operates on standard cloud infrastructure using a single GPU, making it accessible to agencies with varying computational resources. 
While the proposed framework demonstrates strong performance across multiple evaluation settings, certain aspects offer scope for further refinement.
The current evaluation focuses on freeway corridors within a single district; extending validation to urban arterials, mixed-network settings, and multi-city benchmarks would strengthen model generalization. Moreover, federated setup currently uses standard FedAvg aggregation, which does not explicitly account for traffic heterogeneity across clients, and the base model is limited to 1.5B parameters. Exploring traffic-aware aggregation strategies and larger model variants represents a natural extension.

More broadly, this work highlights the potential of combining LLMs with FL for traffic forecasting. While both areas have been studied independently, their integration offers a unified framework that supports privacy preservation, interpretability, and cross-region generalization. Future directions include incorporating multimodal inputs such as incident reports and weather data, extending the framework to hierarchical federated settings, and developing traffic-specific pre-training strategies to reduce local data requirements. Overall, the results indicate that structured prompt design combined with federated training provides a practical approach for developing forecasting systems that are accurate, interpretable, and suitable for deployment in real-world transportation networks.

\section*{Acknowledgment}
During the preparation of this work, the author(s) used Grammarly to enhance the readability and
language of the manuscript. The tool was used solely for grammar and language suggestions. All
content was carefully reviewed and verified by the author.

\bibliographystyle{plainnat}
\bibliography{references}

@article{Wang2012,
author = {Wang, J. and Deng, W. and Zhao, J.},
year = {2012},
month = {01},
pages = {162-167},
title = {Short-term freeway traffic flow prediction based on improved Bayesian combined model},
volume = {42},
journal = {Dongnan Daxue Xuebao (Ziran Kexue Ban)/Journal of Southeast University (Natural Science Edition)},
doi = {10.3969/j.issn.1001-0505.2012.01.030}
}

@InProceedings{Agarap2018,
  author    = {Agarap, Abien Fred M.},
  booktitle = {Proceedings of the 2018 10th International Conference on Machine Learning and Computing},
  title     = {A Neural Network Architecture Combining Gated Recurrent Unit (GRU) and Support Vector Machine (SVM) for Intrusion Detection in Network Traffic Data},
  year      = {2018},
  address   = {New York, NY, USA},
  pages     = {26–30},
  publisher = {Association for Computing Machinery},
  series    = {ICMLC '18},
  doi       = {10.1145/3195106.3195117},
  isbn      = {9781450363532},
  location  = {Macau, China},
  numpages  = {5},
}

@InProceedings{Tang2023,
  author    = {Tang, Huimin and Xue, Nianming and Wang, Gaoli},
  booktitle = {Proceedings of the 2022 10th International Conference on Information Technology: IoT and Smart City},
  title     = {Differentially Private Decentralized Traffic Flow Prediction Approach based on Federated Learning},
  year      = {2023},
  address   = {New York, NY, USA},
  pages     = {280–285},
  publisher = {Association for Computing Machinery},
  series    = {ICIT '22},
  doi       = {10.1145/3582197.3582244},
  isbn      = {9781450397438},
  location  = {Shanghai, China},
  numpages  = {6},
}

@InProceedings{Zhang2024a,
  author    = {Zhang, Yu and Lu, Hua and Liu, Ning and Xu, Yonghui and Li, Qingzhong and Cui, Lizhen},
  booktitle = {Proceedings of the Thirty-Third International Joint Conference on Artificial Intelligence},
  title     = {Personalized federated learning for cross-city traffic prediction},
  year      = {2024},
  series    = {IJCAI '24},
  articleno = {611},
  doi       = {10.24963/ijcai.2024/611},
  isbn      = {978-1-956792-04-1},
  location  = {Jeju, Korea},
  numpages  = {9},
}

@Article{Wei2021,
  author   = {Wei, Wenting and Gu, Huaxi and Li, Baochun},
  journal  = {IEEE Network},
  title    = {Congestion Control: A Renaissance with Machine Learning},
  year     = {2021},
  number   = {4},
  pages    = {262-269},
  volume   = {35},
  doi      = {10.1109/MNET.011.2000603},
}

@Misc{Guo2024,
  author = {Xusen Guo and Qiming Zhang and Junyue Jiang and Mingxing Peng and Meixin Zhu and Hao Yang},
  title         = {Towards Explainable Traffic Flow Prediction with Large Language Models},
  year          = {2024},
  archiveprefix = {arXiv},
  doi           = {10.1016/j.commtr.2024.100150},
  eprint        = {2404.02937},
  primaryclass  = {cs.LG},
  url           = {https://arxiv.org/abs/2404.02937},
}

@Article{kumar2015,
  author    = {Kumar, S.V. and Vanajakshi, L.},
  journal   = {European Transport Research Review},
  title     = {Short-Term Traffic Flow Prediction Using Seasonal ARIMA Model with Limited Input Data},
  year      = {2015},
  number    = {1},
  pages     = {21},
  volume    = {7},
  doi       = {10.1007/s12544-015-0170-8},
  publisher = {Springer},
}

@Misc{Sowinski2016,
  author       = {Piotr Sowi{\'n}ski},
  howpublished = {Zenodo},
  note         = {Licensed under CC-BY-4.0},
  title        = {RiverBench / CityPulse Traffic Dataset (v1.0.3)},
  year         = {2024},
  doi          = {10.5281/zenodo.13525177},
}

@Misc{Liu2024,
  author        = {Ruyang Liu and Chen Li and Haoran Tang and Yixiao Ge and Ying Shan and Ge Li},
  title         = {ST-LLM: Large Language Models Are Effective Temporal Learners},
  year          = {2024},
  archiveprefix = {arXiv},
  eprint        = {2404.00308},
  primaryclass  = {cs.CV},
  doi           = {10.48550/arXiv.2404.00308},
  
}

@Article{Zarza2023,
  author         = {de Zarzà, I. and de Curtò, J. and Roig, Gemma and Calafate, Carlos T.},
  journal        = {Sensors},
  title          = {LLM Multimodal Traffic Accident Forecasting},
  year           = {2023},
  issn           = {1424-8220},
  number         = {22},
  volume         = {23},
  article-number = {9225},
  doi            = {10.3390/s23229225},
  pubmedid       = {38005612},
  ranking        = {rank5},
}

@Misc{Moghadas2024,
  author        = {Seyed Mohamad Moghadas and Bruno Cornelis and Alexandre Alahi and Adrian Munteanu},
  title         = {Strada-LLM: Graph LLM for traffic prediction},
  year          = {2025},
  archiveprefix = {arXiv},
  doi           = {10.48550/arXiv.2410.20856},
  eprint        = {2410.20856},
  primaryclass  = {cs.LG},
  url           = {https://arxiv.org/abs/2410.20856},
}

@Misc{Ren2024,
  author        = {Yilong Ren and Yue Chen and Shuai Liu and Boyue Wang and Haiyang Yu and Zhiyong Cui},
  title         = {TPLLM: A Traffic Prediction Framework Based on Pretrained Large Language Models},
  year          = {2024},
  archiveprefix = {arXiv},
  eprint        = {2403.02221},
  primaryclass  = {cs.LG},
  doi           = {10.48550/arXiv.2403.02221},
}

@Article{Jin2023,
  author  = {Ming Jin and Shiyu Wang and Lintao Ma and Zhixuan Chu and James Y. Zhang and Xiao Long Shi and Pin-Yu Chen and Yuxuan Liang and Yuan-Fang Li and Shirui Pan and Qingsong Wen},
  journal = {ArXiv},
  title   = {Time-LLM: Time Series Forecasting by Reprogramming Large Language Models},
  year    = {2023},
  volume  = {abs/2310.01728},
  doi     = {10.48550/arXiv.2310.01728},
}

@Article{Zhang2020,
  author       = {Zhang, Qi and Chang, Jianlong and Meng, Gaofeng and Xiang, Shiming and Pan, Chunhong},
  journal      = {Proceedings of the AAAI Conference on Artificial Intelligence},
  title        = {Spatio-Temporal Graph Structure Learning for Traffic Forecasting},
  year         = {2020},
  month        = {Apr.},
  number       = {01},
  pages        = {1177-1185},
  volume       = {34},
  abstractnote = {&lt;p&gt;As an indispensable part in Intelligent Traffic System (ITS), the task of traffic forecasting inherently subjects to the following three challenging aspects. First, traffic data are physically associated with road networks, and thus should be formatted as traffic graphs rather than regular grid-like tensors. Second, traffic data render strong spatial dependence, which implies that the nodes in the traffic graphs usually have complex and dynamic relationships between each other. Third, traffic data demonstrate strong temporal dependence, which is crucial for traffic time series modeling. To address these issues, we propose a novel framework named Structure Learning Convolution (SLC) that enables to extend the traditional convolutional neural network (CNN) to graph domains and learn the graph structure for traffic forecasting. Technically, SLC explicitly models the structure information into the convolutional operation. Under this framework, various non-Euclidean CNN methods can be considered as particular instances of our formulation, yielding a flexible mechanism for learning on the graph. Along this technical line, two SLC modules are proposed to capture the global and local structures respectively and they are integrated to construct an end-to-end network for traffic forecasting. Additionally, in this process, Pseudo three Dimensional convolution (P3D) networks are combined with SLC to capture the temporal dependencies in traffic data. Extensively comparative experiments on six real-world datasets demonstrate our proposed approach significantly outperforms the state-of-the-art ones.&lt;/p&gt;},
  doi          = {10.1609/aaai.v34i01.5470}
}

@InProceedings{Li2024,
  author    = {Li, Zhonghang and Xia, Lianghao and Tang, Jiabin and Xu, Yong and Shi, Lei and Xia, Long and Yin, Dawei and Huang, Chao},
  booktitle = {Proceedings of the 30th ACM SIGKDD Conference on Knowledge Discovery and Data Mining},
  title     = {UrbanGPT: Spatio-Temporal Large Language Models},
  year      = {2024},
  address   = {New York, NY, USA},
  pages     = {5351–5362},
  publisher = {Association for Computing Machinery},
  series    = {KDD '24},
  doi       = {10.1145/3637528.3671578},
  isbn      = {9798400704901},
  location  = {Barcelona, Spain},
  numpages  = {12}
}

@Article{Qu2024,
  author  = {Jian Qu and Xiaobo Ma and Jianfeng Li},
  journal = {ArXiv},
  title   = {TrafficGPT: Breaking the Token Barrier for Efficient Long Traffic Analysis and Generation},
  year    = {2024},
  volume  = {abs/2403.05822},
  doi     = {10.48550/arXiv.2403.05822}
}

@Article{Jin2021,
  author   = {KyoHoon Jin and JeongA Wi and EunJu Lee and ShinJin Kang and SooKyun Kim and YoungBin Kim},
  journal  = {Expert Systems with Applications},
  title    = {TrafficBERT: Pre-trained model with large-scale data for long-range traffic flow forecasting},
  year     = {2021},
  issn     = {0957-4174},
  pages    = {115738},
  volume   = {186},
  doi      = {https://doi.org/10.1016/j.eswa.2021.115738},
  ranking  = {rank5}
}

@InProceedings{Gruver2023,
  author    = {Gruver, Nate and Finzi, Marc and Qiu, Shikai and Wilson, Andrew G},
  booktitle = {Advances in Neural Information Processing Systems},
  title     = {Large Language Models Are Zero-Shot Time Series Forecasters},
  year      = {2023},
  editor    = {A. Oh and T. Naumann and A. Globerson and K. Saenko and M. Hardt and S. Levine},
  pages     = {19622--19635},
  publisher = {Curran Associates, Inc.},
  volume    = {36},
  doi       = {10.52202/075280-0861},
  url       = {https://proceedings.neurips.cc/paper_files/paper/2023/file/3eb7ca52e8207697361b2c0fb3926511-Paper-Conference.pdf},
}

@InProceedings{McMahan2016,
  author    = {H. B. McMahan and Eider Moore and Daniel Ramage and Seth Hampson and Blaise Ag{\"u}era y Arcas},
  booktitle = {International Conference on Artificial Intelligence and Statistics},
  title     = {Communication-Efficient Learning of Deep Networks from Decentralized Data},
  year      = {2016},
  doi       = {10.48550/arXiv.1602.05629},
  url       = {https://api.semanticscholar.org/CorpusID:14955348},
}

@Misc{McMahan2016a,
  author = {McMahan, H. and Moore, Eider and Ramage, Daniel},
  month  = {02},
  title  = {Federated Learning of Deep Networks using Model Averaging},
  year   = {2016},
  doi    = {10.48550/arXiv.1602.05629},
  eprint = {1602.05629},
}

@Article{Beutel2020,
  author     = {Daniel J. Beutel and Taner Topal and Akhil Mathur and Xinchi Qiu and Titouan Parcollet and Nicholas D. Lane},
  journal    = {CoRR},
  title      = {Flower: {A} Friendly Federated Learning Research Framework},
  year       = {2020},
  volume     = {abs/2007.14390},
  bibsource  = {dblp computer science bibliography, https://dblp.org},
  eprint     = {2007.14390},
  eprinttype = {arXiv},
  url        = {https://arxiv.org/abs/2007.14390},
}

@Article{Qi2023,
  author    = {Qi, Tao and Chen, Lingqiang and Li, Guanghui and Li, Yijing and Wang, Chenshu},
  journal   = {Applied Soft Computing},
  title     = {FedAGCN: A traffic flow prediction framework based on federated learning and Asynchronous Graph Convolutional Network},
  year      = {2023},
  issn      = {1568-4946},
  month     = may,
  pages     = {110175},
  volume    = {138},
  doi       = {10.1016/j.asoc.2023.110175},
  publisher = {Elsevier BV},
  ranking   = {rank5},
}

@Article{Yurdem2024,
  author   = {Betul Yurdem and Murat Kuzlu and Mehmet Kemal Gullu and Ferhat Ozgur Catak and Maliha Tabassum},
  journal  = {Heliyon},
  title    = {Federated learning: Overview, strategies, applications, tools and future directions},
  year     = {2024},
  issn     = {2405-8440},
  number   = {19},
  pages    = {e38137},
  volume   = {10},
  doi      = {10.1016/j.heliyon.2024.e38137},
}

@Article{Nanayakkara2024,
  author   = {Shanika Iroshi Nanayakkara and Shiva Raj Pokhrel and Gang Li},
  journal  = {Future Generation Computer Systems},
  title    = {Understanding global aggregation and optimization of federated learning},
  year     = {2024},
  issn     = {0167-739X},
  pages    = {114-133},
  volume   = {159},
  doi      = {10.1016/j.future.2024.05.009},
  ranking  = {rank5},
}

@InProceedings{Agarwal2023,
  author    = {Agarwal, Piyush and Sharma, Sachin and Matta, Priya},
  booktitle = {2023 Winter Summit on Smart Computing and Networks (WiSSCoN)},
  title     = {Federated Learning in Intelligent Traffic Management System},
  year      = {2023},
  pages     = {1-6},
  doi       = {10.1109/WiSSCoN56857.2023.10133864},
}

@Article{Bonawitz2019,
  author  = {Keith Bonawitz and Hubert Eichner and Wolfgang Grieskamp and Dzmitry Huba and Alex Ingerman and Vladimir Ivanov and Chlo{\'e} Kiddon and Jakub Konecn{\'y} and Stefano Mazzocchi and H. B. McMahan and Timon Van Overveldt and David Petrou and Daniel Ramage and Jason Roselander},
  journal = {ArXiv},
  title   = {Towards Federated Learning at Scale: System Design},
  year    = {2019},
  volume  = {abs/1902.01046},
  doi     = {10.48550/arXiv.1902.01046},
  url     = {https://api.semanticscholar.org/CorpusID:59599820},
}

@Article{Liu2023,
  author   = {Liu, Lei and Tian, Yuxing and Chakraborty, Chinmay and Feng, Jie and Pei, Qingqi and Zhen, Li and Yu, Keping},
  journal  = {IEEE Transactions on Network and Service Management},
  title    = {Multilevel Federated Learning-Based Intelligent Traffic Flow Forecasting for Transportation Network Management},
  year     = {2023},
  number   = {2},
  pages    = {1446-1458},
  volume   = {20},
  doi      = {10.1109/TNSM.2023.3280515},
}

@Article{Yuan2023,
  author    = {Xiaoming Yuan and Jiahui Chen and Jiayu Yang and Ning Zhang and Tingting Yang and Tao Han and Amir Taherkordi},
  journal   = {IEEE Transactions on Intelligent Transportation Systems},
  title     = {FedSTN: Graph Representation Driven Federated Learning for Edge Computing Enabled Urban Traffic Flow Prediction},
  year      = {2023},
  issn      = {1558-0016},
  pages     = {8738--8748},
  volume    = {24},
  date      = {Aug. 2023},
  doi       = {10.1109/TITS.2022.3157056},
  issue     = {8},
  publisher = {IEEE},
}

@Article{Lu2024,
  author   = {Lu, Zili and Pan, Heng and Dai, Yueyue and Si, Xueming and Zhang, Yan},
  journal  = {IEEE Internet of Things Journal},
  title    = {Federated Learning With Non-IID Data: A Survey},
  year     = {2024},
  number   = {11},
  pages    = {19188-19209},
  volume   = {11},
  doi      = {10.1109/JIOT.2024.3376548},
}

@Article{Reddi2020,
  author     = {Sashank J. Reddi and Zachary Charles and Manzil Zaheer and Zachary Garrett and Keith Rush and Jakub Kone{\v{c}}n{\'y} and Sanjiv Kumar and H. Brendan McMahan},
  journal    = {CoRR},
  title      = {Adaptive Federated Optimization},
  year       = {2020},
  volume     = {abs/2003.00295},
  bibsource  = {dblp computer science bibliography, https://dblp.org},
  doi        = {10.48550/arXiv.2003.00295},
  eprint     = {2003.00295},
  eprinttype = {arXiv},
  url        = {https://arxiv.org/abs/2003.00295},
}

@InProceedings{Karimireddy2019,
  author    = {Sai Praneeth Karimireddy and Satyen Kale and Mehryar Mohri and Sashank J. Reddi and Sebastian U. Stich and Ananda Theertha Suresh},
  booktitle = {International Conference on Machine Learning},
  title     = {SCAFFOLD: Stochastic Controlled Averaging for Federated Learning},
  year      = {2019},
  doi       = {10.48550/arXiv.1910.06378},
  url       = {https://api.semanticscholar.org/CorpusID:214069261},
}

@Article{Zhang2024,
  author   = {Zhang, Hao and Wu, Tingting and Cheng, Siyao and Liu, Jie},
  journal  = {IEEE Internet of Things Journal},
  title    = {CC-FedAvg: Computationally Customized Federated Averaging},
  year     = {2024},
  number   = {3},
  pages    = {4826-4841},
  volume   = {11},
  doi      = {10.1109/JIOT.2023.3300080},
}

@Article{Kairouz2021,
  author     = {Kairouz, Peter and McMahan, H. Brendan and Avent, Brendan and Bellet, Aur\'{e}lien and Bennis, Mehdi and Nitin Bhagoji, Arjun and Bonawitz, Kallista and Charles, Zachary and Cormode, et al.},
  journal    = {Found. Trends Mach. Learn.},
  title      = {Advances and Open Problems in Federated Learning},
  year       = {2021},
  issn       = {1935-8237},
  month      = jun,
  number     = {1–2},
  pages      = {1–210},
  volume     = {14},
  address    = {Hanover, MA, USA},
  doi        = {10.1561/2200000083},
  issue_date = {Jun 2021},
  numpages   = {214},
  publisher  = {Now Publishers Inc.},
  url        = {https://doi.org/10.1561/2200000083},
}

@InProceedings{Yang2024,
  author    = {Yang, Linghua and Chen, Wantong and He, Xiaoxi and Wei, Shuyue and Xu, Yi and Zhou, Zimu and Tong, Yongxin},
  booktitle = {Proceedings of the 30th ACM SIGKDD Conference on Knowledge Discovery and Data Mining},
  title     = {FedGTP: Exploiting Inter-Client Spatial Dependency in Federated Graph-based Traffic Prediction},
  year      = {2024},
  address   = {New York, NY, USA},
  month     = {8},
  pages     = {6105–6116},
  publisher = {Association for Computing Machinery},
  series    = {KDD '24},
  day       = {24},
  doi       = {10.1145/3637528.3671613},
  isbn      = {9798400704901},
  location  = {Barcelona, Spain},
  pagetotal = {12}
}

@article{Feng2024,
author = {Feng, Jian and Du, Cailing and Mu, Qi},
year = {2024},
month = {06},
pages = {210},
title = {Traffic Flow Prediction Based on Federated Learning and Spatio-Temporal Graph Neural Networks},
volume = {13},
journal = {ISPRS International Journal of Geo-Information},
doi = {10.3390/ijgi13060210}
}

@Misc{Touvron2023,
  author        = {Hugo Touvron and Louis Martin and Kevin Stone and Peter Albert and Amjad Almahairi and Yasmine Babaei and Nikolay Bashlykov and Soumya Batra and Prajjwal Bhargava and Shruti Bhosale and Dan Bikel and Lukas Blecher and Cristian Canton Ferrer and Moya Chen and Guillem Cucurull and David Esiobu and Jude Fernandes and Jeremy Fu and Wenyin Fu and Brian Fuller and Cynthia Gao and Vedanuj Goswami and Naman Goyal and Anthony Hartshorn and Saghar Hosseini and Rui Hou and Hakan Inan and Marcin Kardas and Viktor Kerkez and Madian Khabsa and Isabel Kloumann and Artem Korenev and Punit Singh Koura and Marie-Anne Lachaux and Thibaut Lavril and Jenya Lee and Diana Liskovich and Yinghai Lu and Yuning Mao and Xavier Martinet and Todor Mihaylov and Pushkar Mishra and Igor Molybog and Yixin Nie and Andrew Poulton and Jeremy Reizenstein and Rashi Rungta and Kalyan Saladi and Alan Schelten and Ruan Silva and Eric Michael Smith and Ranjan Subramanian and Xiaoqing Ellen Tan and Binh Tang and Ross Taylor and Adina Williams and Jian Xiang Kuan and Puxin Xu and Zheng Yan and Iliyan Zarov and Yuchen Zhang and Angela Fan and Melanie Kambadur and Sharan Narang and Aurelien Rodriguez and Robert Stojnic and Sergey Edunov and Thomas Scialom},
  title         = {Llama 2: Open Foundation and Fine-Tuned Chat Models},
  year          = {2023},
  archiveprefix = {arXiv},
  doi           = {10.48550/arXiv.2307.09288},
  eprint        = {2307.09288},
  primaryclass  = {cs.CL},
  url           = {https://arxiv.org/abs/2307.09288},
}

@Misc{OpenAI2024,
  author        = {OpenAI and others},
  note          = {Full author list available at: \url{https://arxiv.org/abs/2303.08774}},
  title         = {GPT-4 Technical Report},
  year          = {2024},
  archiveprefix = {arXiv},
  doi           = {10.48550/arXiv.2303.08774},
  eprint        = {2303.08774},
  primaryclass  = {cs.CL},
  url           = {https://arxiv.org/abs/2303.08774},
}

@Misc{deepseek,
  author        = {DeepSeek-AI and others},
  note          = {Full author list available at: \url{https://arxiv.org/abs/2412.19437}},
  title         = {DeepSeek-V3 Technical Report},
  year          = {2025},
  archiveprefix = {arXiv},
  doi           = {10.48550/arXiv.2412.19437},
  eprint        = {2412.19437},
  primaryclass  = {cs.CL},
  url           = {https://arxiv.org/abs/2412.19437},
}

@Misc{Grattafiori2024,
  author        = {Aaron Grattafiori and others},
  note          = {Full author list available at: \url{https://arxiv.org/abs/2407.21783}},
  title         = {The Llama 3 Herd of Models},
  year          = {2024},
  archiveprefix = {arXiv},
  doi           = {10.48550/arXiv.2407.21783},
  eprint        = {2407.21783},
  primaryclass  = {cs.AI},
  url           = {https://arxiv.org/abs/2407.21783},
}

@Misc{Liu2023a,
  author        = {Xu Liu and Yutong Xia and Yuxuan Liang and Junfeng Hu and Yiwei Wang and Lei Bai and Chao Huang and Zhenguang Liu and Bryan Hooi and Roger Zimmermann},
  title         = {LargeST: A Benchmark Dataset for Large-Scale Traffic Forecasting},
  year          = {2023},
  archiveprefix = {arXiv},
  doi           = {10.52202/075280-3293},
  eprint        = {2306.08259},
  primaryclass  = {cs.LG},
}

@misc{CaltransWebsite,
  author       = {{California Department of Transportation (Caltrans)}},
  title        = {California Department of Transportation Official Website},
  year         = {2026},
  url          = {https://dot.ca.gov/},
  note         = {Accessed: 2026-03-12}
}

@Misc{Ouyang2022,
  author        = {Long Ouyang and Jeff Wu and Xu Jiang and Diogo Almeida and Carroll L. Wainwright and Pamela Mishkin and Chong Zhang and Sandhini Agarwal and Katarina Slama and Alex Ray and John Schulman and Jacob Hilton and Fraser Kelton and Luke Miller and Maddie Simens and Amanda Askell and Peter Welinder and Paul Christiano and Jan Leike and Ryan Lowe},
  title         = {Training language models to follow instructions with human feedback},
  year          = {2022},
  archiveprefix = {arXiv},
  doi           = {10.52202/068431-2011},
  eprint        = {2203.02155},
  primaryclass  = {cs.CL}
}

@Misc{Qwen2025,
  author        = {Qwen and : and An Yang and Baosong Yang and Beichen Zhang and Binyuan Hui and Bo Zheng and Bowen Yu and Chengyuan Li and Dayiheng Liu and Fei Huang and Haoran Wei and Huan Lin and Jian Yang and Jianhong Tu and Jianwei Zhang and Jianxin Yang and Jiaxi Yang and Jingren Zhou and Junyang Lin and Kai Dang and Keming Lu and Keqin Bao and Kexin Yang and Le Yu and Mei Li and Mingfeng Xue and Pei Zhang and Qin Zhu and Rui Men and Runji Lin and Tianhao Li and Tianyi Tang and Tingyu Xia and Xingzhang Ren and Xuancheng Ren and Yang Fan and Yang Su and Yichang Zhang and Yu Wan and Yuqiong Liu and Zeyu Cui and Zhenru Zhang and Zihan Qiu},
  title         = {Qwen2.5 Technical Report},
  year          = {2025},
  archiveprefix = {arXiv},
  doi           = {10.48550/arXiv.2412.15115},
  eprint        = {2412.15115},
  primaryclass  = {cs.CL},
  url           = {https://arxiv.org/abs/2412.15115},
}

@Misc{unsloth,
  author = {Daniel Han, Michael Han and Unsloth team},
  title = {Unsloth},
  url = {https://github.com/unslothai/unsloth},
  year = {2023}
}

@Misc{Dettmers2023,
  author        = {Tim Dettmers and Artidoro Pagnoni and Ari Holtzman and Luke Zettlemoyer},
  title         = {QLoRA: Efficient Finetuning of Quantized LLMs},
  year          = {2023},
  archiveprefix = {arXiv},
  doi           = {10.52202/075280-0441},
  eprint        = {2305.14314},
  primaryclass  = {cs.LG},
  url           = {https://arxiv.org/abs/2305.14314},
}

@Misc{Hu2021,
  author        = {Edward J. Hu and Yelong Shen and Phillip Wallis and Zeyuan Allen-Zhu and Yuanzhi Li and Shean Wang and Lu Wang and Weizhu Chen},
  title         = {LoRA: Low-Rank Adaptation of Large Language Models},
  year          = {2021},
  archiveprefix = {arXiv},
  doi           = {10.48550/arXiv.2106.09685},
  eprint        = {2106.09685},
  primaryclass  = {cs.CL},
  url           = {https://arxiv.org/abs/2106.09685},
}

@Misc{Cho2014,
  author        = {Kyunghyun Cho and Bart van Merrienboer and Caglar Gulcehre and Dzmitry Bahdanau and Fethi Bougares and Holger Schwenk and Yoshua Bengio},
  title         = {Learning Phrase Representations using RNN Encoder-Decoder for Statistical Machine Translation},
  year          = {2014},
  archiveprefix = {arXiv},
  doi           = {10.3115/v1/d14-1179},
  eprint        = {1406.1078},
  primaryclass  = {cs.CL},
  url           = {https://arxiv.org/abs/1406.1078},
}

@InProceedings{Yu2018,
  author     = {Yu, Bing and Yin, Haoteng and Zhu, Zhanxing},
  booktitle  = {Proceedings of the Twenty-Seventh International Joint Conference on Artificial Intelligence},
  title      = {Spatio-Temporal Graph Convolutional Networks: A Deep Learning Framework for Traffic Forecasting},
  year       = {2018},
  month      = jul,
  pages      = {3634–3640},
  publisher  = {International Joint Conferences on Artificial Intelligence Organization},
  series     = {IJCAI-2018},
  collection = {IJCAI-2018},
  doi        = {10.24963/ijcai.2018/505},
  url        = {http://dx.doi.org/10.24963/ijcai.2018/505},
}

@Misc{Li2018,
  author        = {Yaguang Li and Rose Yu and Cyrus Shahabi and Yan Liu},
  title         = {Diffusion Convolutional Recurrent Neural Network: Data-Driven Traffic Forecasting},
  year          = {2018},
  archiveprefix = {arXiv},
  doi           = {10.48550/arXiv.1707.01926},
  eprint        = {1707.01926},
  primaryclass  = {cs.LG},
  url           = {https://arxiv.org/abs/1707.01926},
}

@InProceedings{Bai2020,
  author    = {Bai, Lei and Yao, Lina and Li, Can and Wang, Xianzhi and Wang, Can},
  booktitle = {Proceedings of the 34th International Conference on Neural Information Processing Systems},
  title     = {Adaptive graph convolutional recurrent network for traffic forecasting},
  year      = {2020},
  address   = {Red Hook, NY, USA},
  publisher = {Curran Associates Inc.},
  series    = {NIPS '20},
  articleno = {1494},
  doi       = {10.48550/arXiv.2007.02842},
  isbn      = {9781713829546},
  location  = {Vancouver, BC, Canada},
  numpages  = {12},
}

@Misc{Sutskever2014,
  author        = {Ilya Sutskever and Oriol Vinyals and Quoc V. Le},
  title         = {Sequence to Sequence Learning with Neural Networks},
  year          = {2014},
  archiveprefix = {arXiv},
  doi           = {10.48550/arXiv.1409.3215},
  eprint        = {1409.3215},
  primaryclass  = {cs.CL},
  url           = {https://arxiv.org/abs/1409.3215},
}

@Article{Guo2022,
  author   = {Guo, Shengnan and Lin, Youfang and Wan, Huaiyu and Li, Xiucheng and Cong, Gao},
  journal  = {IEEE Transactions on Knowledge and Data Engineering},
  title    = {Learning Dynamics and Heterogeneity of Spatial-Temporal Graph Data for Traffic Forecasting},
  year     = {2022},
  number   = {11},
  pages    = {5415-5428},
  volume   = {34},
  doi      = {10.1109/TKDE.2021.3056502},
}

@Article{Li2025,
  author  = {Li, Yuanhui and Mi, Bo and Zeng, Ran},
  journal = {Scientific Reports},
  title   = {FedGDAN: Privacy-preserving traffic flow prediction via federated graph diffusion attention networks},
  year    = {2025},
  month   = {11},
  volume  = {15},
  doi     = {10.1038/s41598-025-24963-z},
}

@Article{Zhang2021,
  author   = {Zhang, Chenhan and Zhang, Shuyu and Yu, James J. Q. and Yu, Shui},
  journal  = {IEEE Transactions on Industrial Informatics},
  title    = {FASTGNN: A Topological Information Protected Federated Learning Approach for Traffic Speed Forecasting},
  year     = {2021},
  number   = {12},
  pages    = {8464-8474},
  volume   = {17},
  doi      = {10.1109/TII.2021.3055283},
}

@Article{Kaur2026,
  author   = {Seerat Kaur and Sukhjit Singh Sehra and Darisuh Ebrahimi and Emad A. Mohammed},
  journal  = {Machine Learning with Applications},
  title    = {A traffic-aware federated learning prediction framework with custom aggregation},
  year     = {2026},
  issn     = {2666-8270},
  pages    = {100829},
  volume   = {23},
  doi      = {https://doi.org/10.1016/j.mlwa.2025.100829},
  url      = {https://www.sciencedirect.com/science/article/pii/S2666827025002129},
}

@Article{Kaur2026a,
  author   = {Seerat Kaur and Sukhjit Singh Sehra and Dariush Ebrahimi and Xin Wang and Jaiteg Singh and Sumeet Kaur Sehra},
  journal  = {Multimodal Transportation},
  title    = {Harnessing Large Language Models for Intelligent Transportation Systems: A Systematic Review},
  year     = {2026},
  issn     = {2772-5863},
  pages    = {100308},
  doi      = {https://doi.org/10.1016/j.multra.2026.100308},
  url      = {https://www.sciencedirect.com/science/article/pii/S2772586326000201},
}

\end{document}